\documentclass[lettersize,journal]{IEEEtran}

\usepackage{amsmath,amsfonts}
\usepackage{bm}
\usepackage[T1]{fontenc}
\usepackage{multirow}
\usepackage[linesnumbered,ruled,vlined]{algorithm2e}
\SetAlFnt{\small}
\SetAlCapFnt{\small}
\SetAlCapNameFnt{\small}
\let\oldnl\nl
\newcommand{\nonl}{\renewcommand{\nl}{\stepcounter{AlgoLine}\let\nl\oldnl}}

\usepackage{xcolor}
\usepackage[table]{xcolor} 
\usepackage{graphicx}

\usepackage{array}
\usepackage{tabularx}
\usepackage{float}
\usepackage{stfloats}

\usepackage{subfigure}
\usepackage[caption=false,font=normalsize,labelfont=sf,textfont=sf]{subfig}
\usepackage{subcaption}
\usepackage{graphicx}
\usepackage{caption}

\usepackage{url}
\usepackage{verbatim}
\usepackage{cite}
\usepackage{CJKutf8}

\SetKwInput{KwIn}{Input}
\SetKwInput{KwOut}{Output}
\SetKwInput{KwRet}{Return}
\DontPrintSemicolon  

\SetKwComment{tcp}{// }{}  
\SetCommentSty{mycommentstyle}

\usepackage{xr}
\makeatletter
\newcommand*{\addFileDependency}[1]{
  \typeout{(#1)}
  \@addtofilelist{#1}
  \IfFileExists{#1}{}{\typeout{No file #1.}}
}
\makeatother

\newcommand*{\myexternaldocument}[1]{
    \externaldocument{#1}
    \addFileDependency{#1.tex}
    \addFileDependency{#1.aux}
}
\myexternaldocument{Supplementary_materials}

\begin{document}

\title{Multi-objective task allocation for electric harvesting robots: a hierarchical route reconstruction approach}

\author{
Peng Chen,
Jing Liang*,~\IEEEmembership{Senior Member,~IEEE},
Hui Song,
Kang-Jia Qiao,
Cai-Tong Yue,~\IEEEmembership{Member,~IEEE},
Kun-Jie Yu,~\IEEEmembership{Member,~IEEE},
Ponnuthurai Nagaratnam Suganthan,~\IEEEmembership{Fellow,~IEEE},
Witold Pedrycz,~\IEEEmembership{Life Fellow,~IEEE}
\thanks{ This work has been submitted to the IEEE for possible publication. Copyright may be transferred without notice, after which this version may no longer be accessible.} 
\thanks{Peng Chen, Kang-Jia Qiao, Cai-Tong Yue, and Kun-Jie Yu are with the School of Electrical and Information Engineering, Zhengzhou University, Zhengzhou 450007, China (e-mail: ty1220899231@163.com; qiaokangjia@yeah.net; yuecaitong@zzu.edu.cn; yukunjie1990@163.com).

Jing Liang is with the School of Electrical Engineering and Automation, Henan Institute of Technology, Xinxiang 453003, China, and also with the School of Electrical and Information Engineering, Zhengzhou University, Zhengzhou 450007, China (e-mail: liangjing@zzu.edu.cn).

Hui Song is with the School of Engineering, RMIT University, Melbourne, VIC, 3000, Australia (email: hui.song@rmit.edu.au).

Ponnuthurai Nagaratnam Suganthan is with the KINDI Center for Computing Research, College of Engineering, Qatar University, Doha, Qatar (e-mail: p.n.suganthan@qu.edu.qa).

Witold Pedrycz is with the Department of Electrical and Computer Engineering, University of Alberta, Edmonton, Canada, and also with the Systems Research Institute of the Polish Academy of Sciences, Warsaw, Poland (e-mail: wpedrycz@ualberta.ca).
}
\thanks{The supplementary materials for this paper are provided via the project website (https://github.com/Peng-ZZU/Supplementary\_materials-for-HRRA.git).} 
}

\markboth{IEEE TRANSACTIONS ON CYBERNETICS}%
{Shell \MakeLowercase{\textit{et al.}}: A Sample Article Using IEEEtran.cls for IEEE Journals}


\maketitle

\begin{abstract}
The increasing labor costs in agriculture have accelerated the adoption of multi-robot systems for orchard harvesting. However, efficiently coordinating these systems is challenging due to the complex interplay between makespan and energy consumption, particularly under practical constraints like load-dependent speed variations and battery limitations. This paper defines the multi-objective agricultural multi-electrical-robot task allocation (AMERTA) problem, which systematically incorporates these often-overlooked real-world constraints. To address this problem, we propose a hybrid hierarchical route reconstruction algorithm (HRRA) that integrates several innovative mechanisms, including a hierarchical encoding structure, a dual-phase initialization method, task sequence optimizers, and specialized route reconstruction operators. Extensive experiments on 45 test instances demonstrate HRRA's superior performance against seven state-of-the-art algorithms. Statistical analysis, including the Wilcoxon signed-rank and Friedman tests, empirically validates HRRA's competitiveness and its unique ability to explore previously inaccessible regions of the solution space. In general, this research contributes to the theoretical understanding of multi-robot coordination by offering a novel problem formulation and an effective algorithm, thereby also providing practical insights for agricultural automation.
\end{abstract}


\section{Introduction}

Escalating global labor expenditures~\cite{li2021many} are driving an irreversible shift toward automated solutions~\cite{wang2022hybrid, yu2021adaptive}. Among various agricultural scenarios, orchard harvesting poses significant challenges to automation~\cite{davidson2016proof} due to its dual requirements for timing and quality. While recent advances in picking robots demonstrate remarkable harvesting capabilities~\cite{xiang2024classification}, single-robot systems are limited in large-scale scenarios. Consequently, deploying and coordinating multiple robots is necessary to achieve higher operational efficiency~\cite{li2020multi}.

The multi-robot task allocation (MRTA) problem~\cite{xiong2022probability} comprises two main components in agricultural settings: route construction and route-robot assignment. The former defines the sequence of task nodes within a single trip, while the latter determines the overall task distribution among robots. Existing work shows that random or approximate task allocation, which neglects task characteristics, leads to system-wide inefficiencies~\cite{yan2023load}. This makes optimized task allocation a critical area of research.

In agricultural management operations, MRTA faces conflicting objectives between maximal completion time (makespan) and energy consumption~\cite{dai2023multi}: minimizing makespan benefits from parallel harvesting with frequent returns, while energy minimization encourages full-load returns to reduce trip frequency. This fundamental conflict, coupled with the strong NP-hard nature of makespan minimization~\cite{stewart2023optimising}, makes most existing MRTA methods unsuitable for direct applications~\cite{guo2024effective}.

Current agricultural automation research demonstrates progress in harvesting~\cite{dai2023multi}, spraying~\cite{dong2024effective}, and weeding systems~\cite{wang2024multi}. Nevertheless, these studies often simplify or overlook crucial factors such as the dynamic interplay of load, speed, energy, and battery management ~\cite{dorling2016vehicle,mcnulty2022review} within multi-trip harvesting scenarios. Addressing these multifaceted constraints simultaneously presents a significant but not well addressed challenge in agricultural robotics. These characteristics significantly expand and complicate the search space~\cite{montoya2017electric}. To distinguish this unique problem from existing agricultural MRTA problems, we define it as the agricultural multi-electrical-robot task allocation (AMERTA) problem. This new formulation specifically characterizes the operational constraints and complexity found in orchard environments.

To address these challenges, it is essential to design targeted solution approaches. Exact algorithms are not the preferred choice due to their complexity in obtaining optimal solutions within specified time constraints~\cite{choudhury2022dynamic}, poor performance on large-scale problems~\cite{xia2024two}, and limitations in handling multi-objective optimization. Instead, heuristic methods are more suitable as the primary solution approach. However, considering that exact methods can quickly locate optimal solutions for small-scale single-objective problems, mixed integer linear programming (MILP) models are specifically formulated to handle route allocation in this research. Therefore, a hybrid algorithm called hierarchical route reconstruction algorithm (HRRA) is proposed to solve the AMERTA problem. The main contributions of this research include:

\begin{itemize}
    \item Formulation of a mathematical model for the AMERTA problem that captures the dynamics of payload-dependent robot speed, energy consumption patterns, and battery capacity constraints under practical orchard conditions;
    \item Development of the HRRA, which incorporates a hierarchical solution encoding structure, a variable load-limit dual-phase initialization method, two distinct optimization mechanisms for intra-route and inter-route sequences, as well as charging-based and split-based route reconstruction mechanisms;
    \item Design and implementation of comprehensive experimental studies through a newly constructed benchmark set of 45 test instances with varying problem scales. Extensive computational results demonstrate HRRA's superior performance against seven representative algorithms. 
\end{itemize}

This paper is structured as follows: a systematic review of relevant literature is presented in Section~\ref{Literature review}. The AMERTA problem formulation and mathematical model are established in Section~\ref{Problem description and modeling}. Section~\ref{proposed algorithm} elaborates the proposed HRRA methodology. Comprehensive experimental validation and performance analysis are provided in Section~\ref{experimental results}. Finally, Section~\ref{conclusion} concludes with key findings and future works.

\section{Literature review}
\label{Literature review}
This study investigates the AMERTA problem, which is situated at the intersection of several key research domains. It fundamentally integrates principles from the electric vehicle routing problem (EVRP) with the broad field of MRTA. To provide a comprehensive background, this review first discusses literature from EVRP, which contributes critical energy-related aspects such as battery capacity constraints. We then survey general MRTA approaches that provide foundational frameworks for routing and assignment, before finally focusing on the specific context of agricultural MRTA.

\subsection{EVRP research}
The widespread adoption of electric vehicles (EVs) has been driven by recent advances in new energy technologies~\cite{cano2018batteries}. A key challenge in EV operations is the need to monitor battery capacity alongside load constraints~\cite{amiri2023robust}. To address this issue, existing works have developed diverse optimization strategies, including variable neighborhood search~\cite{yilmaz2022variable}, artificial bee colony (ABC)~\cite{guo2023low}, and ant colony optimization (ACO)~\cite{fan2024two}.

Traditional EVRP studies have largely relied on simplified assumptions of constant energy consumption rates between locations~\cite{comert2023new}. More recent research has incorporated non-linear functions to better reflect real-world conditions, introducing enhanced algorithms such as improved particle swarm genetic hybridization~\cite{bruglieri2023matheuristic}, adaptive genetic algorithms~\cite{li2020electric}, and bi-strategy optimization~\cite{lu2019bi}. However, these models have inadequately addressed the unique characteristics of orchard transportation operations, where the dynamic nature of harvesting loads influences both energy consumption and operational velocity, presenting optimization challenges beyond conventional EVRP scenarios.

Charging strategy optimization has represented a crucial component of EVRP research. Traditional EVRP models have typically involved multiple charging stations and single delivery trips~\cite{xia2024two}. To enhance charging flexibility, various charging mechanisms, including partial charging~\cite{xiao2024electric}, battery swapping~\cite{wu2021survey}, and mobile charging stations~\cite{beyazit2023electric}, have been explored. In comparison, robots in the orchard harvesting context must make multiple depot visits for unloading, thus battery replacement at the depot offers advantages in infrastructure cost and routing efficiency. However, this has introduced operational complexities: tasks may require premature termination due to power constraints, and load updates due to battery replacement impact subsequent task scheduling.

Traditional EVRP studies have predominantly concentrated on single-objective optimization, while the limited research addressing multiple objectives often resorts to weighted-sum approaches that transform multi-objective problems into single-objective ones~\cite{amiri2023robust,comert2023new}. This simplified treatment makes existing methods difficult to directly apply to the AMERTA problem.

\subsection{General MRTA research}

In MRTA problems, factors such as finite robot capacity and number necessitate inter-robot task allocation. This assignment conceptually aligns with the generalized assignment problem (GAP)~\cite{song2022solving} and its extensions~\cite{fathollahi2023efficient,zhou2022hybrid}. However, ensuring operational efficiency in practical MRTA scenarios also critically involves detailed task scheduling for each robot to optimize route construction and the sequence of tasks within individual trips~\cite{chen2025reinforced}. Furthermore, the inherent need to simultaneously optimize multiple conflicting objectives significantly elevates the complexity of the MRTA problems~\cite{zhang2022research,qian2024cooperative}.

To resolve these challenges, the field has seen a rise in sophisticated multi-objective optimization algorithms. For instance, Xue et al. \cite{xue2021hybrid} introduced a hybrid competitive optimization algorithm with adaptive grid partitioning to handle large-scale, many-objective MRTA problems. Similarly, Wei et al. \cite{wei2020particle} developed a multi-objective particle swarm optimization that refines the Pareto front using a probability-based leader selection strategy. Other notable advancements include the work of Zhang et al. \cite{zhang2022research}, who integrated the Lin–Kernighan–Helsgaun heuristic to pre-generate high-quality solutions for multi-objective evolutionary algorithms (MOEAs). More recently, Wen et al. \cite{wen2024indicator} proposed an indicator-based MOEA with a hybrid encoding scheme.

While these methods are powerful, the multi-trip nature of agricultural harvesting, combined with its unique operational demands (such as load-dependent travel times and opportunistic battery management) necessitates novel algorithmic solutions that go beyond the scope of existing MRTA approaches.

\subsection{Agricultural MRTA research}

Agricultural MRTA addresses the coordination of multiple robots in agricultural scenarios, necessitating the consideration of specific constraints tailored to the operational characteristics. Dai et al.~\cite{dai2023multi} made the first attempt by developing a multi-objective discrete ABC (MODABC) algorithm for harvesting robot coordination, benchmarking it against adapted versions of classical algorithms like NSGA-\uppercase\expandafter{\romannumeral2}~\cite{deb2002fast} and MOEA/D~\cite{zhang2007moea}. Inspired by this work, Guo et al.~\cite{guo2024effective} proposed a collaborative discrete ABC (CDABC) algorithm featuring multiple neighborhood structures and a dynamic neighborhood strategy to balance global exploration and local exploitation. For spraying operations, Dong et al.~\cite{dong2024effective} developed an effective multi-objective evolutionary algorithm (AMOEA) that uniquely combines non-dominated solution information for global exploration with iterative greedy strategies for local refinement. In optimizing multi-weeding robot assignments, Kang et al.~\cite{kang2024multi} introduced a multi-objective teaching-learning-based optimization (MOTLBO) algorithm incorporating heuristic initialization methods and a multi-teacher framework. In addition, the scope has further expanded to multi-type robot cooperation, as demonstrated by Wang et al.~\cite{wang2024multi} in coordinating weeding robots with spraying drones.

Despite demonstrated efficacy in constrained scenarios, these population-based approaches have shared common shortcomings: their fixed-dimension solution representation has constrained modeling flexibility. Additionally, all their operations are typically performed on global task sequences, which has restricted the ability to effectively optimize individual trips. Furthermore, the absence of battery replacement strategies in these approaches has made them inadequate for task assignment for electric robotic systems. 

In contrast, the proposed HRRA is specifically engineered to address these limitations by integrating both population-based and individual-based optimization approaches~\cite{jia2022confidence}, employing a hierarchical solution encoding structure that enables the individual representation of each solution. This structure allows for variable-dimensionality global sequences among different solutions, thereby enhancing solution flexibility in manipulation. Furthermore, each route and the entire set of routes assigned to an individual robot can be optimized independently, significantly improving the flexibility of optimization. Critically, to handle the core constraints of AMERTA, two reconstruction mechanisms are proposed: a charging-based approach to address battery capacity constraints and a split-based method to handle load capacity limitations.

\section{Problem description and modeling}
\label{Problem description and modeling}
\subsection{Problem description}

\begin{figure}[htp]
    \centering
    \includegraphics[width=8.8cm]{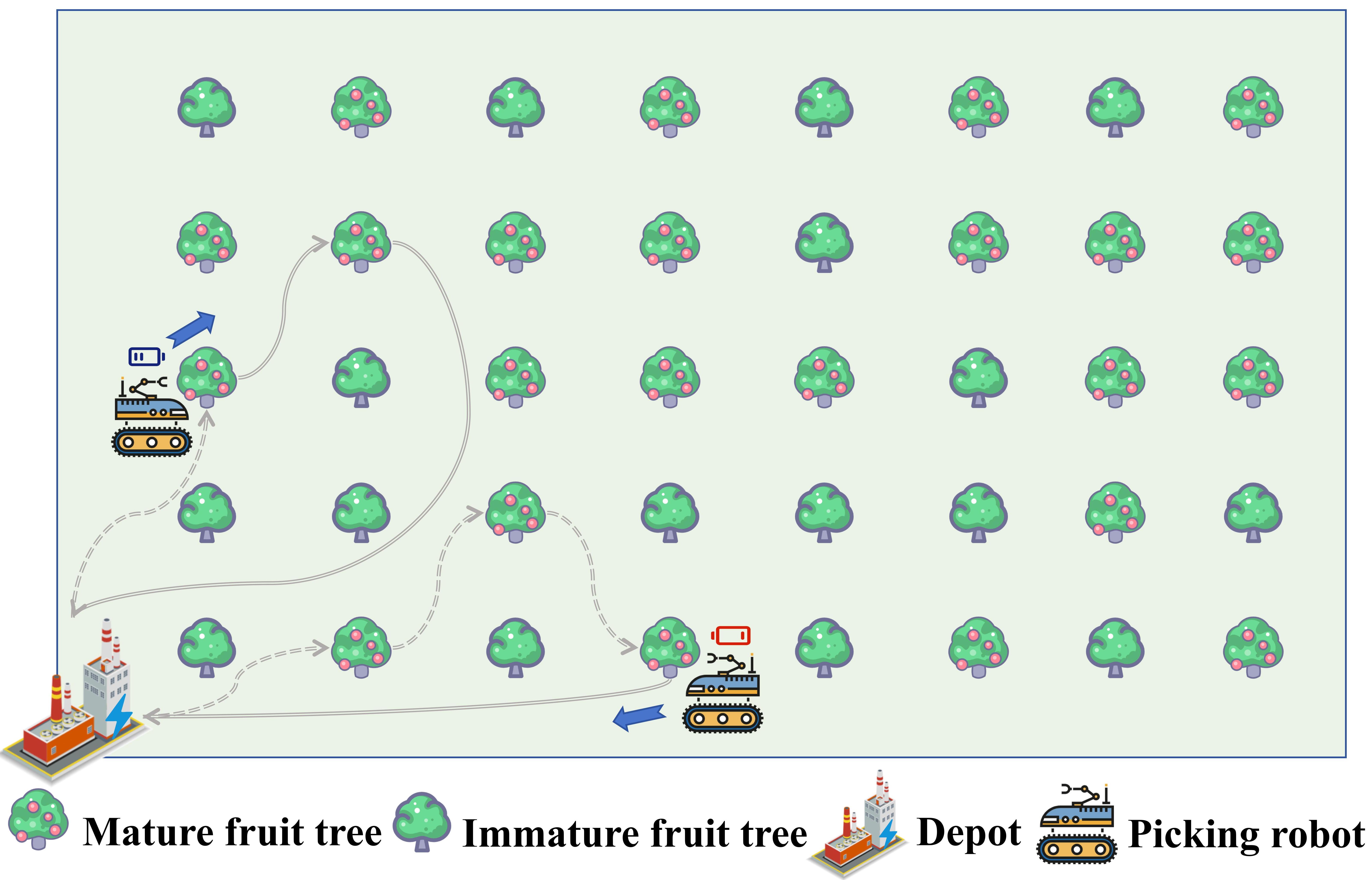}
    \caption{Schematic diagram of orchard scene}
    \label{fruit trees}
\end{figure}

Consider an orchard with uniformly planted trees as shown in~Fig.~\ref{fruit trees}, where trees with ripe fruits exceeding a maturity threshold are designated as task nodes, while others serve as obstacles. The orchard contains $n$ task nodes with different yields. All fruits must be harvested to maintain product quality. The physical layout assumes known coordinates for all task nodes. Travel distances between any two nodes are pre-calculated, representing the shortest navigable paths within this orchard environment.
\vspace{-0.7em}
\vspace{-0.7em}

Initially, $r$ identical picking robots, fully charged, are stationed at the depot. Each task node is assigned to a single robot. Furthermore, to ensure both operational clarity and the efficiency of harvesting, the complete task at any given node is performed by the assigned robot in a single visit. Due to capacity constraints, robots must make multiple trips to complete their assigned tasks. To save resources, battery replacement at the depot is only permitted when charge level falls below a threshold ($B_\textrm{th}$), except for cases where power depletion coincides with task completion. Robots always leave the depot empty-loaded.

This study aims to simultaneously minimize both the makespan ($T_\textrm{max}$) and the total energy consumption ($E_\textrm{total}$) of all robots.

\subsection{Problem modeling}
\textbf{Sets and parameters}
\begin{flalign}
& N = \{0,1,\ldots,n\}: \text{ set of nodes (0 represents depot)} && \notag \\[3pt]
& R = \{1,\ldots,r\}: \text{ set of robots}&& \notag \\[3pt]
& S = \{1,\ldots,s\}: \text{ set of all possible routes} && \notag \\[3pt]
& S^r: \text{ complete route of robot } r && \notag \\[3pt]
& d_{ij}: \text{ distance between nodes } i \text{ and } j && \notag \\[3pt]
& q_i: \text{ fruit yield at node } i  && \notag \\[3pt]
& Q=300: \text{ robot load capacity \cite{yu2020frequency}} && \notag \\[3pt]
& W= 100: \text{ empty robot weight} && \notag \\[3pt]
& B = 432: \text{ battery capacity \cite{mcnulty2022review}} && \notag \\[3pt]
& B_\textrm{th} = 0.2B: \text{ battery threshold for replacement \cite{mitici2022electric}} && \notag \\[3pt]
& g = 9.81: \text{ gravitational acceleration \cite{eckert2012a+}} && \notag \\[3pt]
& \mu = 0.05: \text{ rolling resistance coefficient \cite{valero2017influence}} && \notag \\[3pt]
& \eta = 0.8: \text{ energy efficiency coefficient \cite{DOE2024EV}} && \notag \\[3pt]
& e = 0.5: \text{ unit picking energy \cite{lou2024analysis}} && \notag
\end{flalign}
\begin{flalign}
& \tau = 7: \text{ unit picking time \cite{xiong2019development} } && \notag \\[3pt]
& P_\textrm{max} \approx 3.9: \text{ maximum power output \cite{singh2022design}} && \notag \\[3pt]
& E_{ij}: \text{ energy consumption from node } i \text{ to node } j && \notag \\[3pt]
& E^s_i: \text{ picking energy consumption at node } i && \notag \\[3pt]
& T_{ij}: \text{ travel time from node } i \text{ to node } j && \notag \\[3pt]
& T^s_i: \text{ picking time at node } i && \notag \\[3pt]
& t_\textrm{swap} = 150: \text{ time to replace a battery \cite{yang2015battery}} && \notag \\[3pt]
& T^b_i: \text{battery replacement time after finishing task } i && \notag \\[3pt]
& E_\textrm{total}: \text{ total energy consumption} && \notag \\[3pt]
& T_\textrm{max}: \text{ maximum completion time, makespan} &&\notag\\[3pt]
& n_{s}:  \text{ the last node in route $S^r$} && \notag
\end{flalign}

\textbf{Decision variables}
\begin{flalign}
&\begin{aligned}
& x_{ij} = \begin{cases}
1, & \text{if robot travels from node } i \text{ to node } j \\
0, & \text{otherwise}
\end{cases} \\
& \qquad \forall i,j \in N, i \neq j \\
& \parbox[t]{0.8\columnwidth}{\small Defines the robot's path between nodes}
\end{aligned} && \notag \\[3ex]
&\begin{aligned}
& y_i = \begin{cases}
1, & \text{if battery is replaced after task } i \\
0, & \text{otherwise}
\end{cases} \\
& \qquad \forall i \in N \\
& \parbox[t]{0.8\columnwidth}{\small Determines if a battery swap occurs after node $i$}
\end{aligned} && \notag \\[3ex]
&\begin{aligned}
& L_i \geq 0  \qquad \forall i \in N \\
& \parbox[t]{0.8\columnwidth}{\small Tracks the cumulative load of the robot upon departing from node $i$}
\end{aligned} && \notag \\[3ex]
&\begin{aligned}
& b_i \geq 0 \qquad \forall i \in N \\
& \parbox[t]{0.8\columnwidth}{\small Represents the remaining battery energy level after completing the task at node $i$}
\end{aligned} && \notag \\[3ex]
&\begin{aligned}
& z_{rs} = \begin{cases}
1, & \text{if robot } r \text{ executes route } s \\
0, & \text{otherwise}
\end{cases} \\
& \qquad \forall r \in R, s \in S \\
& \parbox[t]{0.8\columnwidth}{\small Assigns a complete route $s$ to a specific robot $r$}
\end{aligned} && \notag
\end{flalign}

\textbf{Energy and time components}
\begin{flalign}
&\begin{aligned}
& E_{ij} = \frac{d_{ij}(W + L_i)g\mu}{\eta} \times 10^{-3} \qquad \forall i,j \in N, i \neq j \\
& \parbox[t]{0.8\columnwidth}{\small Calculates the travel energy, which is dependent on the distance and the robot's current load $L_i$}
\end{aligned} && \notag \\[3ex]
&\begin{aligned}
& E^s_i = \begin{cases}
eq_i, & i \in N \setminus \{0\} \\
0, & i = 0
\end{cases} \qquad \forall i \in N \\
& \parbox[t]{0.8\columnwidth}{\small Calculates the energy consumed for the picking operation at a task node}
\end{aligned} && \notag 
\end{flalign}
\begin{flalign}
&\begin{aligned}
& T_{ij} = \frac{E_{ij}}{P_\textrm{max}} \qquad \forall i,j \in N, i \neq j \\
& \parbox[t]{0.8\columnwidth}{\small Determines the travel time based on the travel energy and the robot's maximum power output}
\end{aligned} && \notag \\[3ex]
&\begin{aligned}
& T^s_i = \begin{cases}
\tau q_i, & i \in N \setminus \{0\} \\
0, & i = 0
\end{cases} \qquad \forall i \in N \\
& \parbox[t]{0.8\columnwidth}{\small The time required for the picking operation, proportional to the yield}
\end{aligned} && \notag \\[3ex]
&\begin{aligned}
& T^b_i = y_i t_\textrm{swap} \qquad \forall i \in N \setminus \{0\} \\
& \parbox[t]{0.8\columnwidth}{\small Represents the time penalty incurred if a battery swap is performed}
\end{aligned} && \notag
\end{flalign}

\textbf{Objective functions}
\begin{flalign}
&\begin{aligned}
& \min E_\textrm{total} = \sum_{r \in R} \sum_{(i,j) \in S^r} (E_{ij} + E^s_j) \\
& \parbox[t]{0.8\columnwidth}{\small Minimizes the total energy consumption}
\end{aligned} && \notag \\[3ex]
&\begin{aligned}
& \min T_\textrm{max} = \max_{r \in R} \sum_{(i,j) \in S^r} (T_{ij} + T^s_j + T^b_j) \\
& \parbox[t]{0.8\columnwidth}{\small Minimizes the maximum completion time (makespan)}
\end{aligned} && \notag
\end{flalign}

\textbf{Constraints}
\begin{flalign}
&\begin{aligned}
& \sum_{j \in N, j \neq i} x_{ij} = \sum_{j \in N, j \neq i} x_{ji} \qquad \forall i \in N \\
& \parbox[t]{0.8\columnwidth}{\small Maintains route feasibility through flow conservation}
\end{aligned} && \notag \\[3ex]
&\begin{aligned}
& L_0 = 0 \\
& \parbox[t]{0.8\columnwidth}{\small Ensures zero load whenever robots depart from the depot}
\end{aligned} && \notag \\[3ex]
&\begin{aligned}
& L_j =  \sum_{i \in N \setminus \{j\}} (L_i + q_j)x_{ij} \qquad \forall j \in N \setminus \{0\} \\
& \parbox[t]{0.8\columnwidth}{\small Tracks load changes considering inter-node transfers}
\end{aligned} && \notag \\[3ex]
&\begin{aligned}
& L_i \leq Q \qquad \forall i \in N \\
& \parbox[t]{0.8\columnwidth}{\small Prevents overloading at any node}
\end{aligned} && \notag \\[3ex]
&\begin{aligned}
& b_i - E_{ij} - E^s_j \geq 0 \qquad \forall i,j \in N, i \neq j \\
& \parbox[t]{0.8\columnwidth}{\small Ensures energy feasibility for movements and services}
\end{aligned} && \notag \\[3ex]
&\begin{aligned}
& y_i = \begin{cases}
1, & \text{if } b_i \leq B_\textrm{th}  \land i \neq n_s \\
0, & \text{otherwise}
\end{cases} \qquad \forall i \in N \\
& \parbox[t]{0.8\columnwidth}{\small Manages battery replacement decisions}
\end{aligned} && \notag 
\end{flalign}
\begin{flalign}
&\begin{aligned}
& b_i = \begin{cases}
B, & \text{if } y_i = 1 \\
b_{i-1} - E_{i-1,i}(L_{i-1}) - E^s_i, & \text{otherwise}
\end{cases} \\
& \parbox[t]{0.8\columnwidth}{\small Updates battery energy considering consumption and replacement}
\end{aligned} && \notag \\[3ex]
&\begin{aligned}
& \sum_{r \in R} z_{rs} = 1 \qquad \forall s \in S \\
& \parbox[t]{0.8\columnwidth}{\small Ensures proper route-robot assignment}
\end{aligned} && \notag
\end{flalign}
\textbf{Where:}
\begin{itemize}
    \item All time-related units are in seconds, all distance-related units are in meters, all energy-related units are in kilojoules, all weight-related units are in kilograms, and all power-related units are in kilowatts;
    \item All model parameters are set based on existing research or practical scenario considerations~\cite{dai2023multi}, solely for the purpose of numerical simulation testing of the algorithms.
\end{itemize}

\section{Proposed algorithm}
\label{proposed algorithm}
\subsection{Solution representation}
\label{subsec_solution representation}
\begin{figure}[htp]
    \centering
    \includegraphics[width=9cm]{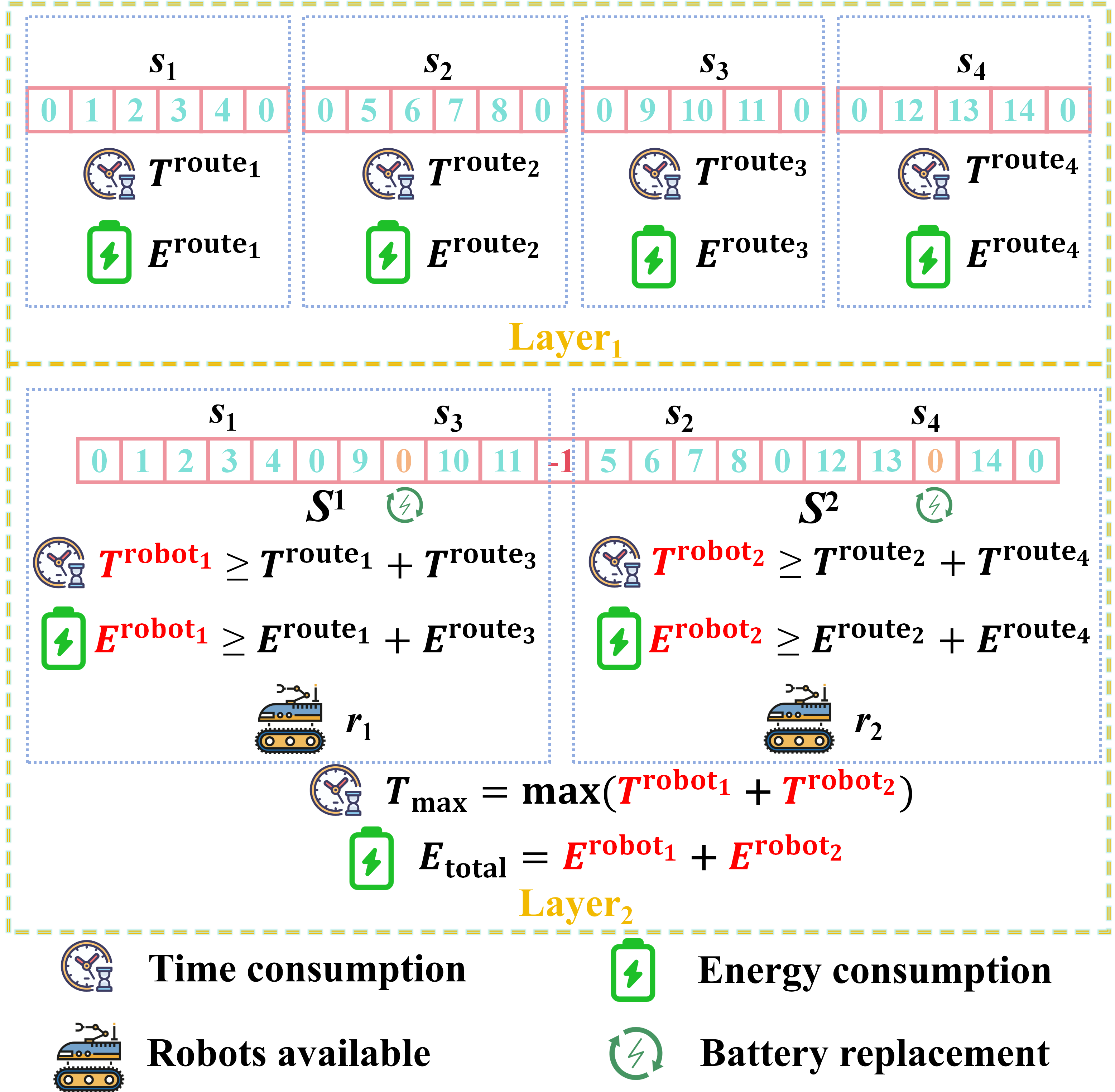}
    \caption{Solution representation}
    \label{Solution representation.jpg}
\end{figure}

This study proposes a hierarchical solution encoding structure that effectively captures the complex characteristics of route construction and route-robot assignment through multi-level information organization, as illustrated in Fig.~\ref{Solution representation.jpg}. For simplicity, tasks are directly represented by their indices, and sequences indicate task execution order. The encoding scheme comprises two organically connected layers: the micro-route layer (layer$_1$) and the macro-scheduling layer (layer$_2$).

In layer$_1$, each independent task execution route is encoded as a triplet $\{S_i, T^{\textrm{route}_i}, E^{\textrm{route}_i}\}$, where $S_i$ represents the complete task sequence including depot nodes (node $0$), while $T^{\textrm{route}_i}$ and $E^{\textrm{route}_i}$ denote the execution time and energy consumption of the $i$-th route, respectively. This design enables independent evaluation and optimization of each route's performance metrics, providing reliable decision support for upper-level route allocation.

Layer$_2$ constructs solutions as a multi-dimensional structure with the following key components:
\begin{itemize}
    \item Global task sequence: employs `-1' as robot task separators and `$0$' as intra-robot route separators, achieving compact task allocation representation;
    \item Robot-task mapping sequence: records complete task sequence $S^r$ for each robot $r$;
    \item Performance metrics set: includes cumulative energy consumption $E^\textrm{robot}_r$ and total completion time $T^\textrm{robot}_r$ for each robot;
    \item Charging position record: maintains indices of all charging points (corresponding to yellow elements `0' in Fig.~\ref{Solution representation.jpg}) during task execution, facilitating subsequent route optimization.
\end{itemize}

This bi-level encoding structure offers several distinct advantages, including of:
\begin{itemize}
    \item Decoupling route construction and route-robot assignment representation, reducing problem complexity;
    \item Adopting microscopic path-level representation and evaluation metrics, enabling path-specific local optimization while significantly improving solution assessment efficiency by avoiding redundant calculations—only the optimized path's metrics need updating, leaving other unchanged paths' evaluations intact;
    \item Facilitating task adjustments between different robots and routes through a compact global task sequence design using separators;
    \item Maintaining separate robot-specific task sequences, energy consumption, and time metrics at the layer$_2$ to address battery constraints, as actual execution sequences cannot be simply combined from layer$_1$ routes.
\end{itemize}

Compared to traditional linear sequence representations~\cite{jia2022confidence}, this hierarchical encoding structure maintains solution completeness and interpretability while significantly enhancing computational efficiency and optimization performance. This innovative representation approach provides new research directions for solving multi-robot collaborative task planning problems.

\subsection{Variable load-limit dual-phase initialization}

Initial population quality and diversity significantly influence algorithm convergence and solution quality. The proposed variable load-limit dual-phase initialization method (VLDIM) comprises two key phases: route construction and route-robot assignment.

\subsubsection{Route construction}
\begin{figure}[htp]
    \centering
    \includegraphics[width=9cm]{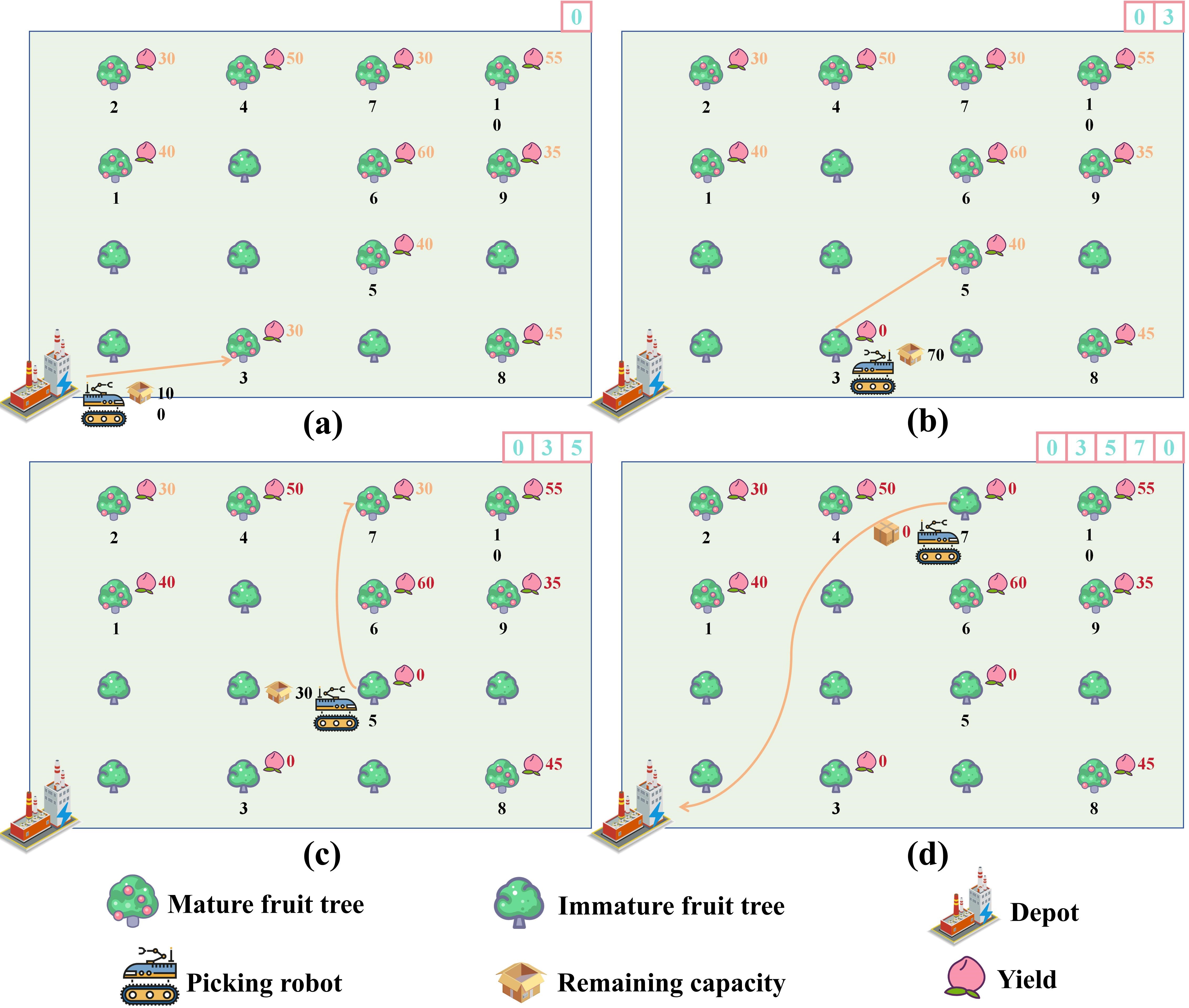}
    \caption{Route construction}
    \label{route construction.jpg}
\end{figure}

The route construction phase employs a distance-based greedy strategy to build task sequences. For tasks in set $N$, the algorithm first selects the nearest node to the robot's initial position as the first task node. Subsequently, it iteratively selects the nearest unvisited node to the current task as the next destination, as illustrated in Fig.~\ref{route construction.jpg}. For instance, task nodes 3 and 5 are sequentially selected based on proximity, followed by task node 7 as the next ideal node under load constraints. After executing these tasks, the robot returns to the depot, completing an initial route. Following this principle, remaining tasks are organized into corresponding routes to complete the layer$_1$ of the solution.

To account for the impact of real-time load on robot velocity, we introduce a linear load-limit strategy. For the $p$-th solution in the population, its load limit $Q_p$ is calculated as:
\begin{equation}
\label{Q_p}
    Q_p = Q \cdot (1-\frac{1-\theta}{pnum} \cdot p)
\end{equation}
where $Q$ denotes the maximum vehicle capacity, $\theta$ represents the limit parameter, and $pnum$ is population size. This linearly decreasing load limit design ensures solution feasibility while enhancing population diversity through varied route sequence lengths.

\begin{figure}[htp]
    \centering
    \includegraphics[width=9cm]{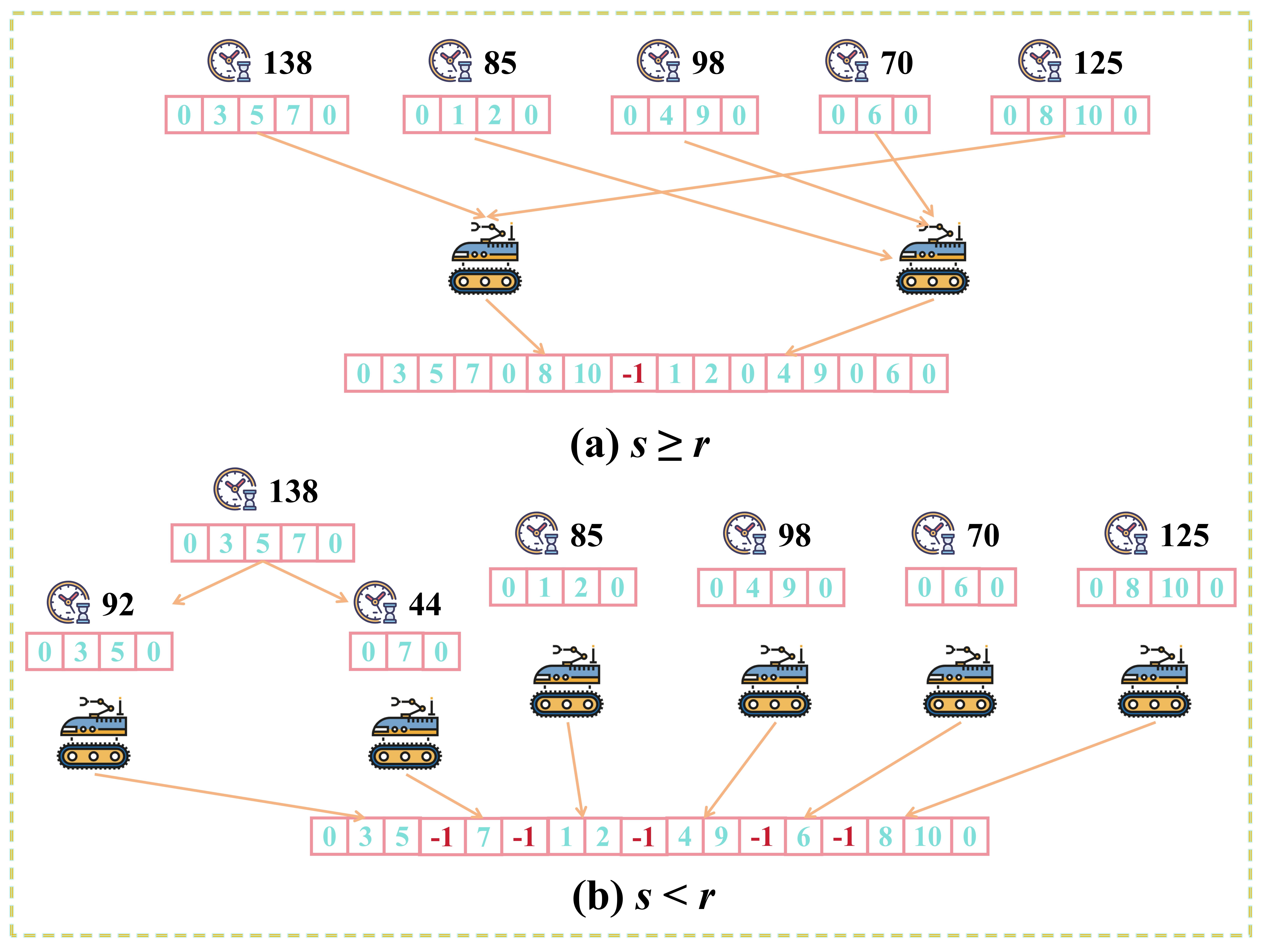}
    \caption{Route-robot assignment}
    \label{route assignment.jpg}
\end{figure}

\subsubsection{Route-robot assignment}
After layer$_1$ route construction, routes must be efficiently allocated to robots to construct the layer$_2$, as shown in Fig.~\ref{route assignment.jpg}. When route count $s$ equals or exceeds robot count $r$, the following MILP$_1$ model is employed:

\textbf{Sets and input parameters:}
\begin{itemize}\setlength{\itemsep}{3pt}
    \item $S = \{S_1, S_2, \ldots, S_s\}$: The set of $s$ pre-constructed routes from layer$_1$.
    \item $R = \{R_1, R_2, \ldots, R_r\}$: The set of $r$ available identical robots.
\end{itemize}

\vspace{0.2cm}
\begin{subequations}\label{eq:MILP1}
\textbf{Decision variables:}
\begin{itemize}\setlength{\itemsep}{3pt}
        \item $\begin{aligned}
        &z_{ij} = \begin{cases}
        1, & \text{if route $S_i$ is assigned to robot } j \\
        0, & \text{otherwise}
        \end{cases} \\[0.1cm]
        & \qquad \forall i \in \{1,\ldots,s\}, j \in \{1,\ldots,r\}
        \end{aligned}$
    \item $C_j$: completion time of robot $j$
    \item $C_\textrm{max}$: makespan, the objective of the research
\end{itemize}

\vspace{0.3cm}
\textbf{Objective function:}
\begin{equation}
    \min C_\textrm{max}
    \label{eq:MILP1_o1}
\end{equation}

\vspace{0.2cm}
\textbf{Constraints:}
\begin{align}
    &\sum_{j=1}^r z_{ij} = 1, && \forall i \in \{1,\ldots,s\} \label{eq:MILP1_c1}
\end{align}
\begin{align} 
    &C_j = \sum_{i=1}^s T^\textrm{route}_i \cdot z_{ij}, && \forall j \in \{1,\ldots,r\} \label{eq:MILP1_c2}\\[0.2cm]
    &C_\textrm{max} \geq C_j, && \forall j \in \{1,\ldots,r\} \label{eq:MILP1_c3}
\end{align}
\end{subequations}
where $T^\textrm{route}_i$ represents $S_i$'s execution time. The constraint~(\ref{eq:MILP1_c1}) ensures each route's assignment, constraint~(\ref{eq:MILP1_c2}) calculates robot completion times, and constraint~(\ref{eq:MILP1_c3}) defines maximum makespan. This model achieves balanced route distribution by minimizing objective~(\ref{eq:MILP1_o1}).

When route count is less than robot count, the longest routes are split into two time-balanced sub-routes iteratively until reaching the robot count. This assignment strategy ensures balanced task distribution while providing quality initial solutions for subsequent optimization.

\subsection{Task sequence optimization}

\subsubsection{Distance-based route reordering mechanism for intra-route optimization}

The initialization phase merely clusters tasks within each route at the layer$_1$, without guaranteeing the optimality of execution sequences. The distance-based route reordering mechanism (DRRM) aims to enhance solution quality by optimizing task execution order within individual routes. This mechanism not only improves individual route performance but also minimizes efficiency losses during route merging due to energy constraints. Specifically, the proposed optimization strategy comprises two key components: distance-based reordering and 2-opt local search~\cite{uddin2023improvement}.

Initially, the algorithm computes the distance between each task node and the depot, then reorders the task sequence in descending order of these distances. This strategy is based on the rationale that prioritizing distant tasks reduces the adverse effects of real-time loads while completing these energy-intensive tasks when battery levels are sufficient, thereby minimizing unnecessary charging operations.

\begin{figure}[htp]
    \centering
    \includegraphics[width=9cm]{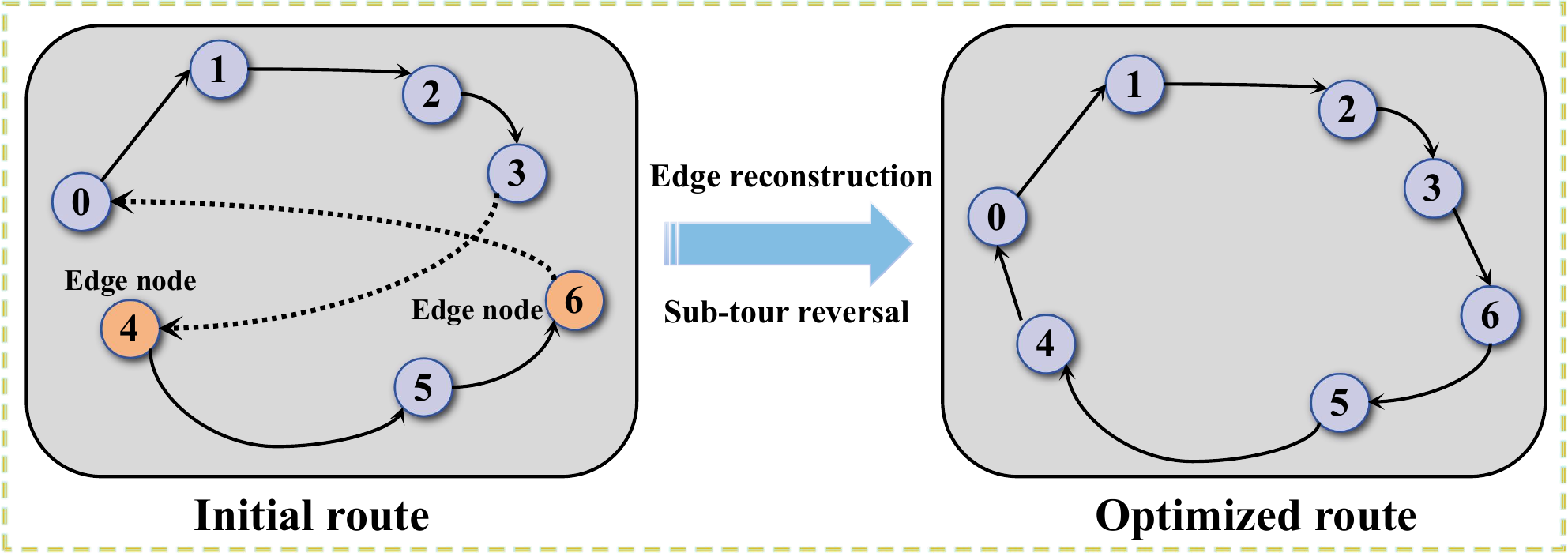}
    \caption{2-opt operation}
    \label{fig:2opt}
\end{figure}

Following distance reordering, the algorithm applies 2-opt local search for fine-grained sequence adjustment. As illustrated in Fig.~\ref{fig:2opt}, this method systematically explores neighborhood solutions by exchanging a pair of connection nodes of positions $i$ and $j$ and reversing the subsequence between remaining connection nodes. New sequences are accepted if they demonstrate superior performance (lower energy consumption or shorter execution time).

\begin{figure}[htp]
    \centering
    \includegraphics[width=9cm]{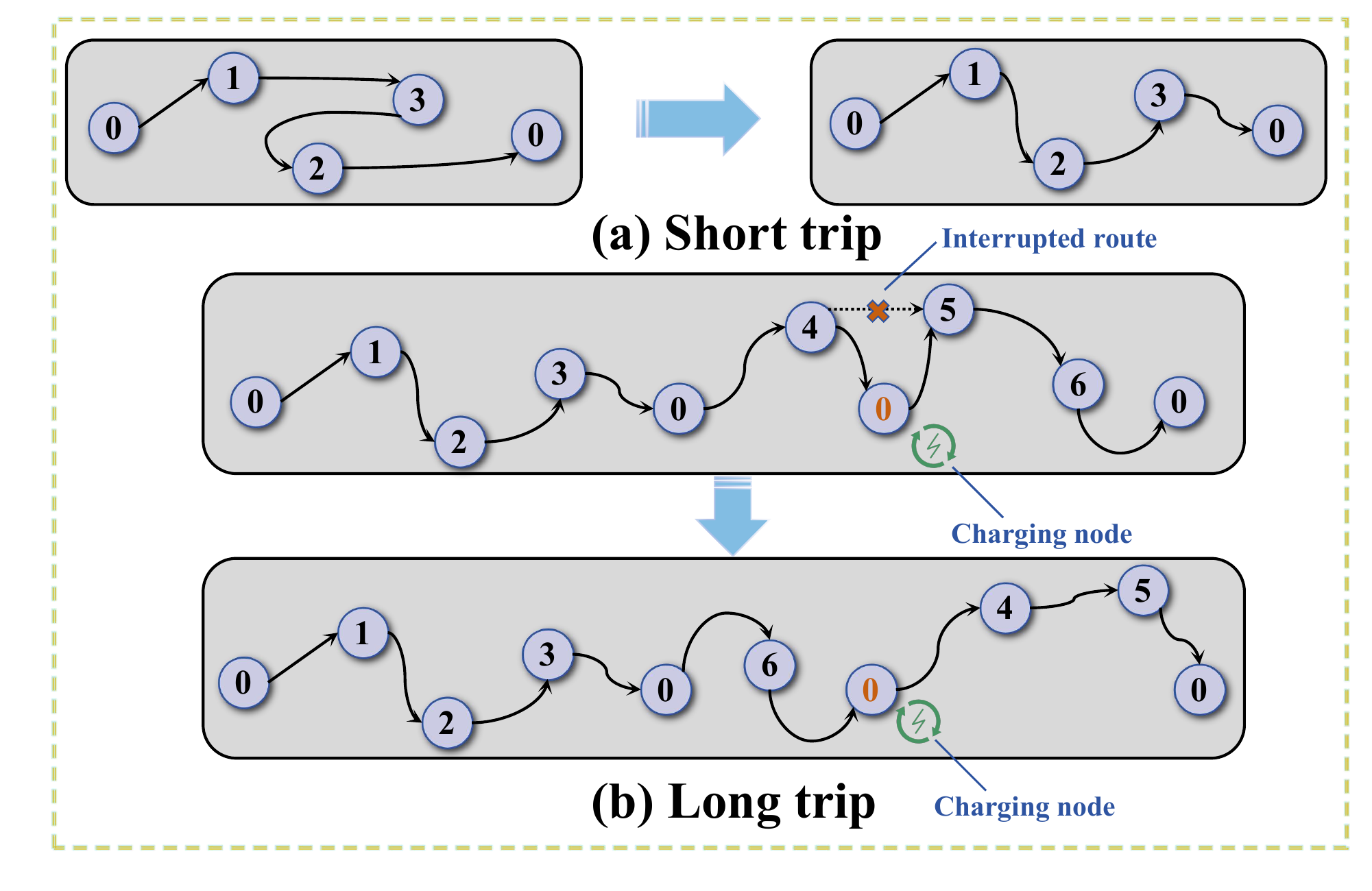}
    \caption{Intra-route optimization}
    \label{fig:intra_opt}
\end{figure}

As shown in Fig.~\ref{fig:intra_opt}(a), intra-route optimization enhances route rationality by eliminating unnecessary detours through task resequencing. More significantly, as shown in Fig.~\ref{fig:intra_opt}(b), intra-route optimization mitigates the influence of structural disruptions by adjusting task sequences in merged routes, thereby improving overall solution quality. This optimization mechanism enhances individual route efficiency while creating essential operational prerequisites for subsequent route merging and assignment operations.

\subsubsection{Task-based route redistribution mechanism for inter-route optimization}

To further enhance solution quality, a task-based route redistribution mechanism (TRRM) is proposed. This mechanism optimizes task allocation structures among robots through task exchange and task reallocation operations for each non-dominated solution in the population.

Specifically, TRRM executes the following operations with equal probability~\cite{dai2023multi}:
\begin{itemize}
    \item Task exchange: randomly selects route sequences from two robots and exchanges task nodes between them;
    \item Task reallocation: randomly selects two robots and redistributes tasks from the robot with longer completion time to the shorter one.
\end{itemize}

\begin{figure}[htp]
    \centering
    \includegraphics[width=9cm]{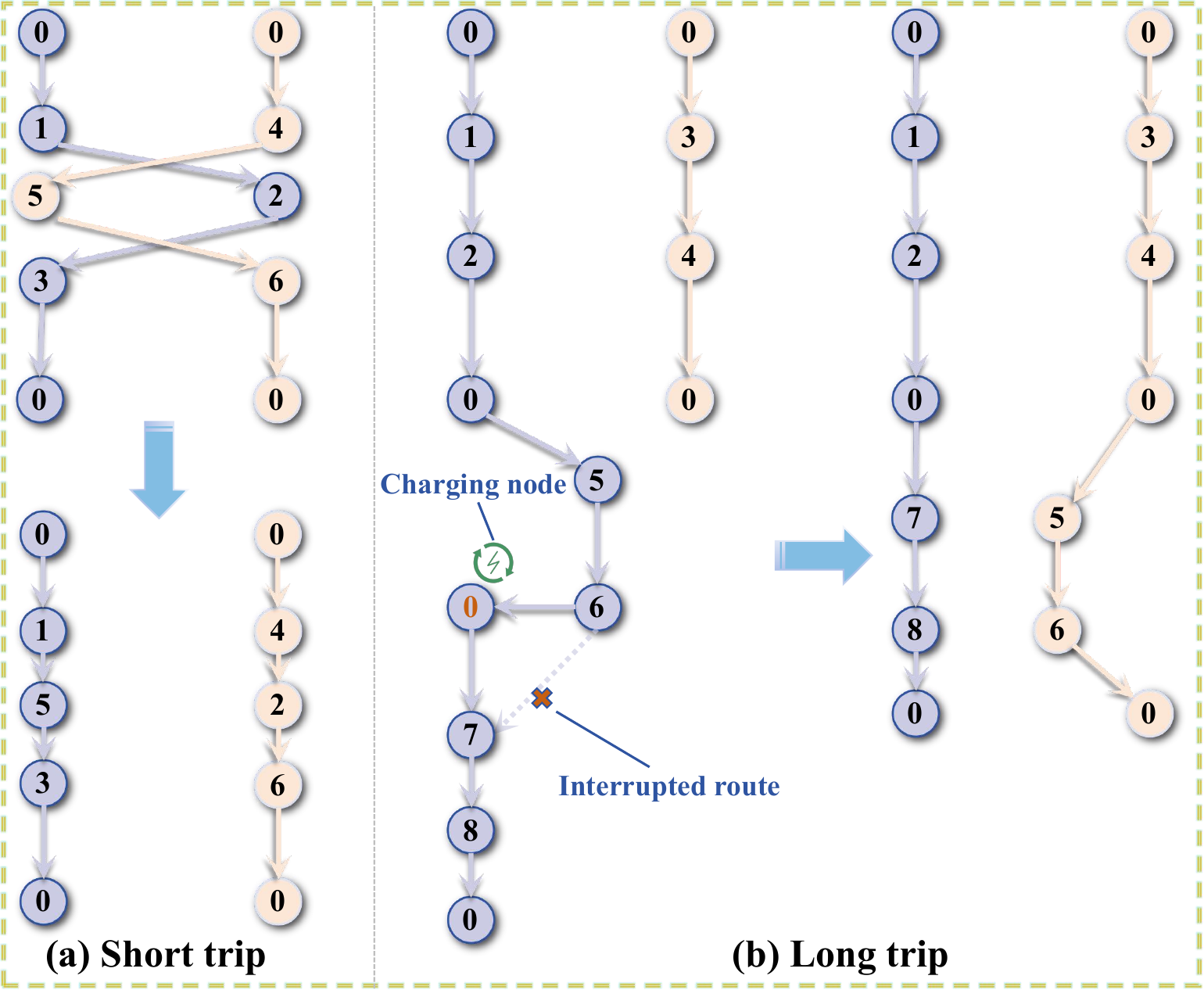}
    \caption{Inter-route optimization}
    \label{fig:inter_opt}
\end{figure}

As illustrated in Fig.~\ref{fig:inter_opt}(a), TRRM's primary function is to optimize each robot's route structure through task redistribution. More importantly, as shown in Fig.~\ref{fig:inter_opt}(b), TRRM optimizes battery energy utilization efficiency through flexible task adjustments. This optimization mechanism not only reduces unnecessary charging operations but also improves battery energy efficiency while ensuring task completion. This optimization mechanism provides more efficient execution plans for multi-robot systems by balancing task allocation and energy utilization.

\subsection{Charging-based route reconstruction}

To further optimize the impact of battery capacity on route structures, a charging-based route reconstruction mechanism (CRRM) is proposed. For each non-dominated solution, this mechanism first extracts task sequences following the last charging operation of each robot (TLC) from the layer$_2$, reorganizes and optimizes these tasks, then optimally redistributes them through a MILP$_2$ model.

CRRM comprises three key steps:
\begin{itemize}
    \item Task extraction: extracting TLC while preserving pre-charging sequences. For robots without charging history, all tasks are extracted. The mechanism terminates if no robot has performed charging operations;
    \item Sequence optimization: applying DRRM to the extracted task set to obtain optimized execution sequences;
    \item Task redistribution: employing the following MILP$_2$ model to reassign optimized task sequences.
\end{itemize}

\begin{subequations}\label{eq:MILP2}
\textbf{Decision variables:}
\begin{itemize}\setlength{\itemsep}{3pt}
    \item $\begin{aligned}
        &z_{ij} = \begin{cases}
        1, & \text{if task } i \text{ is assigned to robot } j \\
        0, & \text{otherwise}
        \end{cases} \\[0.1cm]
        & \qquad \forall i \in \{1,\ldots,n\}, j \in \{1,\ldots,r\}
        \end{aligned}$
    \item $\begin{aligned}
        &w_j = \begin{cases}
        1, & \text{if robot } j \text{ receives new tasks} \\
        0, & \text{otherwise}
        \end{cases} \\[0.1cm]
        & \qquad \forall j \in \{1,\ldots,r\}\\[0.1cm]
        \end{aligned}$
    \item $C_\textrm{max}$: makespan
\end{itemize}
\vspace{0.3cm}
\textbf{Objective function:}
\begin{equation}
    \min C_\textrm{max}
    \label{eq:MILP2_o1}
\end{equation}

\vspace{0.2cm}
\textbf{Constraints:}
\begin{align}
    &\sum_{j=1}^r z_{ij} = 1, \qquad \qquad \forall i \in \{1,\ldots,n\} \label{eq:MILP2_c1}\\[0.2cm]
    &\sum_{i=1}^n z_{ij} \leq n w_j, \quad \quad \, \forall j \in \{1,\ldots,r\} \label{eq:MILP2_c2}
\end{align}
\begin{align}
    &\sum_{i=1}^n T^\textrm{route}_i z_{ij} + t_\textrm{swap} w_j +  T^\textrm{init}_j  \leq C_\textrm{max} \label{eq:MILP2_c3} \\
    &\qquad  \forall j \in \{1,\ldots,r\}  \nonumber 
\end{align}
\end{subequations}
where $t_\textrm{swap}$ denotes battery replacement time and $T^\textrm{init}_j$ represents robot $j$'s execution time of pre-charging sequence. The constraint~(\ref{eq:MILP2_c1}) ensures task assignment completeness, constraint~(\ref{eq:MILP2_c2}) defines robot utilization status, constraint~(\ref{eq:MILP2_c3}) calculates and limits the makespan. This model achieves balanced task redistribution by minimizing objective~(\ref{eq:MILP2_o1}).

\begin{figure}[htp]
    \centering
    \includegraphics[width=9cm]{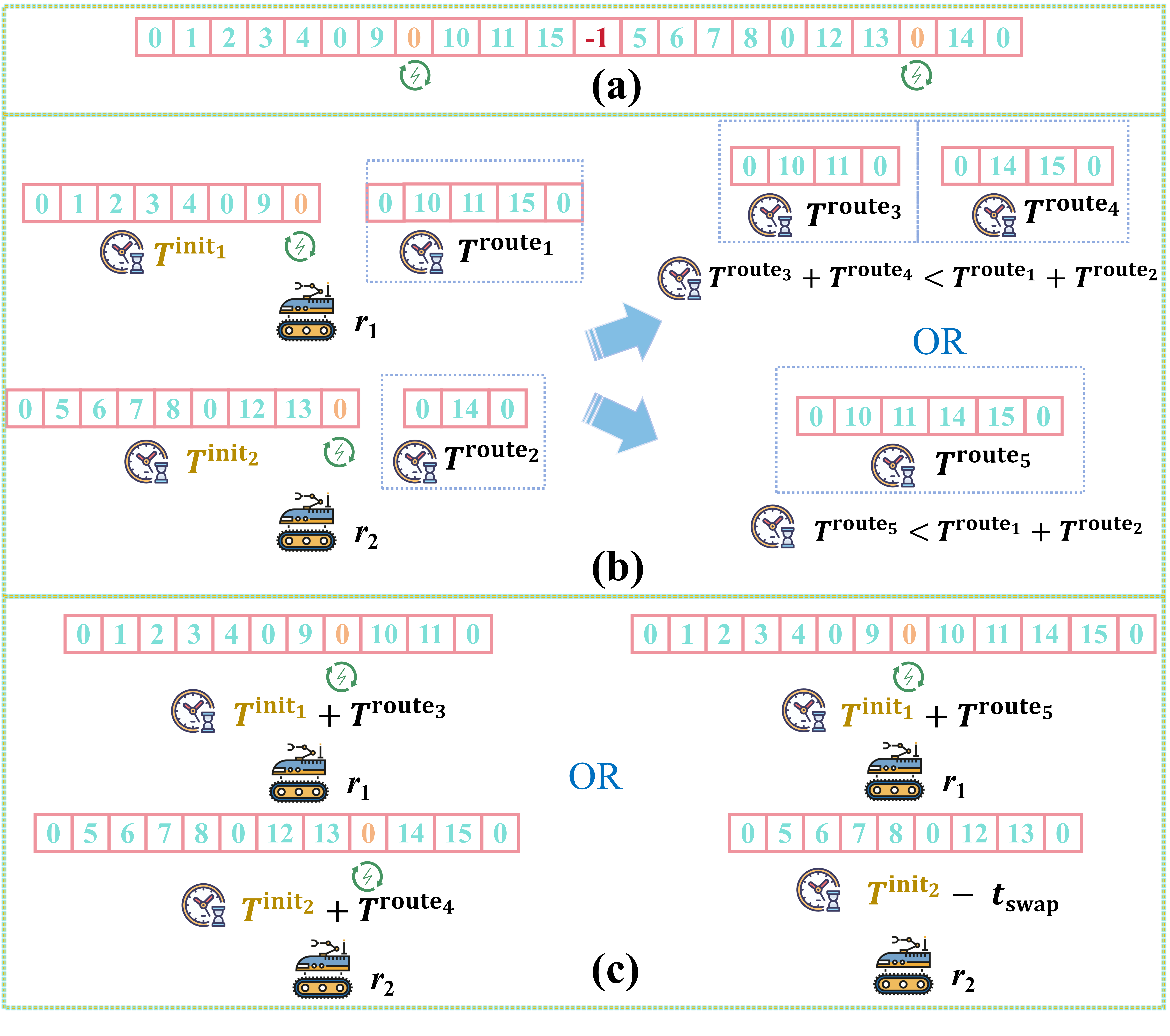}
    \caption{Charging-based route reconstruction}
    \label{fig:CRRM}
\end{figure}

As illustrated in Fig.~\ref{fig:CRRM}(a), we consider a global task sequence awaiting optimization, scheduled for execution by two robots. TLC are reorganized, potentially yielding multiple reconstruction patterns due to conflicting evaluation metrics, as shown in Fig.~\ref{fig:CRRM}(b). Finally, reconstructed routes are reassigned to robots while minimizing makespan (allowing some robots to remain unassigned for new routes, thereby avoiding extra charging operations), as depicted in Fig.~\ref{fig:CRRM}(c).

CRRM primarily enhances overall execution efficiency through last-charging task sequence reconstruction. This optimization mechanism not only weakens the impact of structural disruptions caused by charging operations but also achieves more balanced task allocation while maintaining battery energy constraints.

\subsection{Split-based route reconstruction}

To further optimize the balance of task allocation, a split-based route reconstruction mechanism (SRRM) is proposed. This mechanism achieves dynamic load balancing by iteratively identifying and splitting the most time-consuming routes, followed by a comprehensive reallocation of all routes.

The SRRM comprises three key steps:
\begin{itemize}
    \item Route identification: restore the global route sequence from the layer$_2$ of the solution to the layer$_1$, and identify the route with the longest execution time;
    \item Route splitting: divide the longest route into two sub-routes with approximately equal execution times;
    \item Route reallocation: redistribute all routes to robots using MILP$_1$.
\end{itemize}

In the route splitting phase, the algorithm employs an iterative greedy strategy: starting from either end of the longest route (randomly selecting the sequence head or tail), it progressively transfers task nodes (excluding depot) to the new route until finding the optimal splitting point that minimizes the execution time difference between the two sub-routes.

\begin{figure}[htp]
    \centering
    \includegraphics[width=9cm]{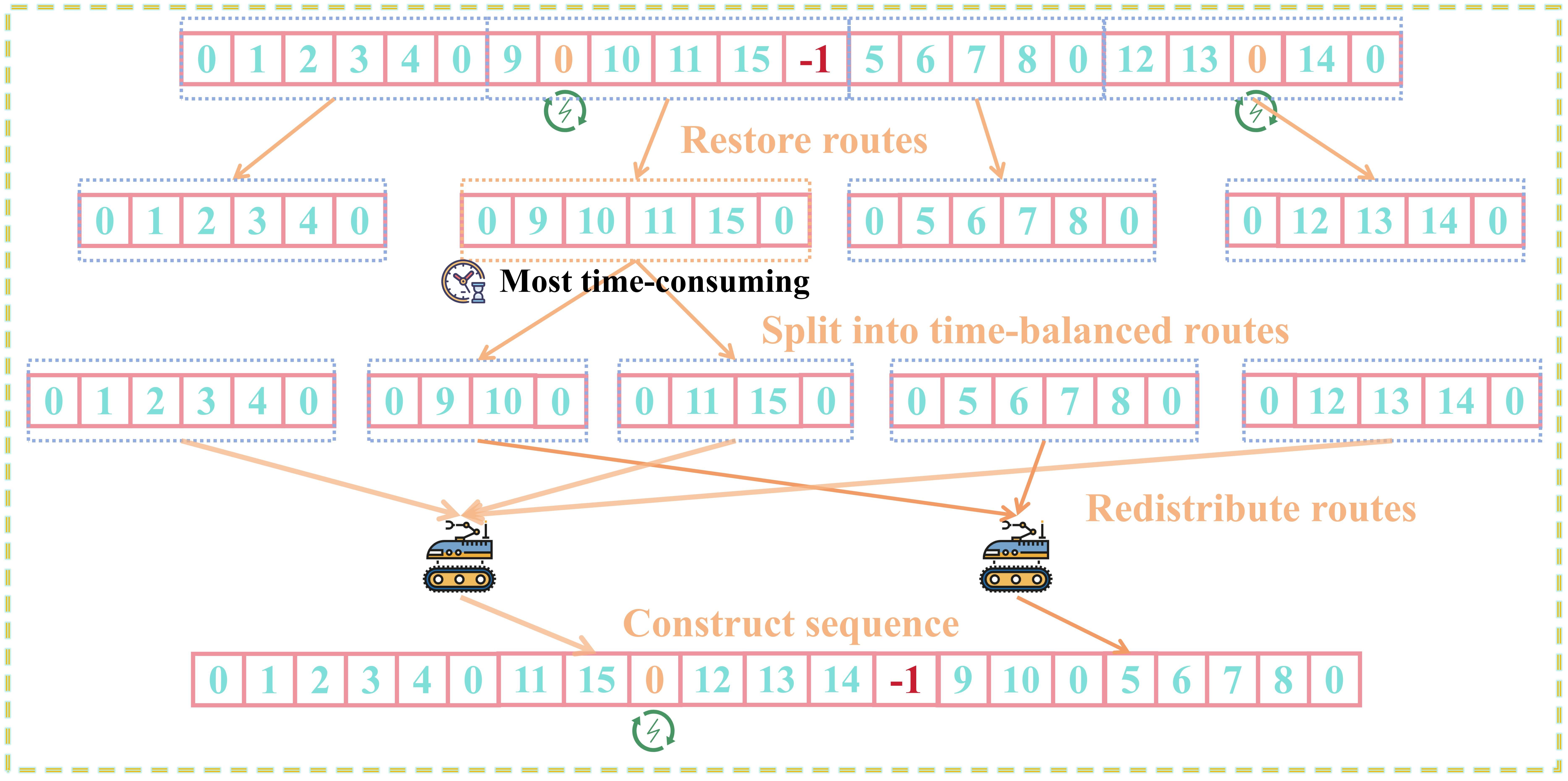}
    \caption{Split-based route reconstruction mechanism}
    \label{fig:SRRM}
\end{figure}

The complete process is illustrated in Fig.~\ref{fig:SRRM}. The primary function of SRRM is to optimize temporal balance of tasks through dynamic route splitting and reconstruction, which facilitates makespan optimization by providing more flexible task allocation options.

\begin{figure*}[htp]
    \centering
    \includegraphics[width=18cm]{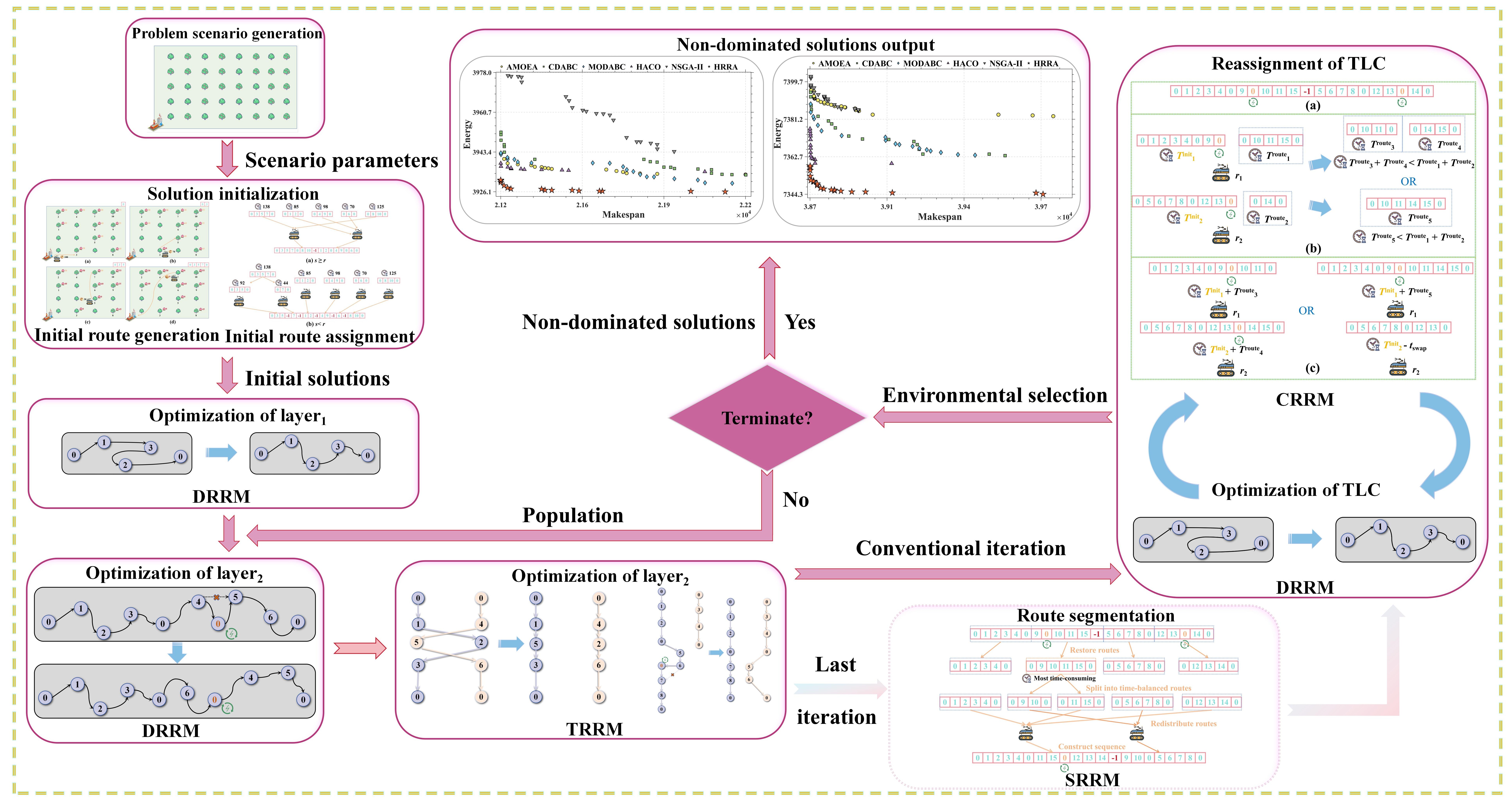}
    \caption{Framework of HRRA}
    \label{fig:framework}
\end{figure*}

\begin{algorithm}[htb]
\SetAlgoLined
\KwIn{%
    \begin{tabular}[t]{ll}
        Task set: & $N$\\
        Number of robots: & $r$\\
        Population size: & $pnum$\\
        Time limit: & $time\_limit$\\
        Problem parameters: & $params$
    \end{tabular}
}
\vspace{0.1em}
\KwOut{Set of non-dominated solutions}
\vspace{0.1em}
$\text{CurrentTime} \leftarrow 0$\;
\vspace{0.1em}
\tcp{Initialization with Algorithm 2}
$P \leftarrow \text{VLDIM}(N, r, pnum, params)$\;
\vspace{0.1em}
$obj \leftarrow \text{Evaluate}(P)$\;
\vspace{0.1em}
\tcp{Initial route optimization with Algorithm 3}
Optimize routes in layer$_1$ using DRRM\;
\vspace{0.1em}
$iter \leftarrow 0$\;
\vspace{0.1em}
\tcp{Main optimization loop}
\While{$\text{CurrentTime} < time\_limit$}{  
\vspace{0.1em}
\tcp{Algorithm 3}
    Optimize $S^r$ in layer$_2$ of each solution in $P$ using DRRM\;
    \vspace{0.1em}
    \tcp{Algorithm 4}
    $(P', obj) \leftarrow \text{TRRM}(P, pnum, r, obj, params)$\;
    \vspace{0.1em}
    $F_1 \leftarrow \text{NonDominatedSort}(obj)$\;
    \vspace{0.1em}
    \tcp{Algorithm 5}
    $(P'') \leftarrow \text{CRRM}(P'(F_1), params)$\;
    \vspace{0.1em}
    $iter \leftarrow iter +1$\;
    \vspace{0.1em}
    $iter\_time \leftarrow \text{CurrentTime} \div iter$\;
    \vspace{0.1em}
    \tcp{Final refinement with Algorithm 6}
    \If{$\text{CurrentTime} + iter\_time < time\_limit$}{
    \vspace{0.1em}
        $(P''') \leftarrow \text{SRRM}(P''(F_1), params)$\;
    }
    \vspace{0.1em}
    $P \leftarrow \text{EnvironmentalSelection}(P''' \cup P)$\;
}
\vspace{0.1em}
$F_1 \leftarrow \text{NonDominatedSort}(obj)$\;
\vspace{0.1em}
\Return{$P(F_1)$}
\caption{Multi-objective optimization framework}
\label{alg:framework}
\end{algorithm}

\subsection{Complete flow of HRRA}

The proposed algorithm integrates multiple reconstruction mechanisms within a hierarchical optimization framework, as illustrated in Fig.~\ref{fig:framework}. Following scenario initialization, a population of solutions are initialized using VLDIM with scenario-specific parameters. The initial routes in layer$_1$ are then optimized through DRRM. Subsequently, the algorithm iteratively performs the following procedures until termination criteria are met: $S^r$ in layer$_2$ are first optimized using DRRM, followed by TRRM optimization of global task sequences. CRRM and DRRM are then applied to extract and optimize TLC from each non-dominated solution, after which these optimized TLC are redistributed among robots. The algorithm records the average computational time per iteration and, when the remaining time is insufficient for another complete iteration, executes SRRM for final solutions refinement. In each iteration, environmental selection is applied to filter for a high-quality population~\cite{deb2002fast}. Upon termination, the algorithm outputs a set of trade-off solutions that balance multiple objectives.

This process is summarized in Algorithm~\ref{alg:framework}, where the corresponding Algorithms~\ref{alg:initialization}--\ref{alg:SRRM} for each component are presented in Section~\ref{S-Algorithms} of the supplementary materials due to the space limitation. Additionally, comprehensive complexity analysis of the algorithm is provided in Section~\ref{S-complexity}.

\section{Experimental studies and analysis}
\label{experimental results}
This section presents a comprehensive evaluation of the proposed algorithm through systematic comparative experiments. The experimental setup, including benchmark problems, performance metrics, and experimental environment, is first described in Section~\ref{Experimental setup}. Afterwards, parameter sensitivity analysis for $\theta$ is conducted in Section \ref{S-parameter} of the supplementary materials, revealing optimal performance at $\theta = 0.8736$. The effectiveness of each algorithmic component is then validated through ablation studies in Section \ref{S-ablation}. Subsequently, comparative results against state-of-the-art algorithms are presented and analyzed in Section~\ref{sec:experiments }. And statistical analysis with confidence intervals is presented in Section~\ref{Confidence interval}. Ultimately, the performance comparison of the default outputs from various algorithms is validated in Section \ref{S-default outputs} when no specific preferences are held by decision-makers.

\begin{table}[h]
\centering
\caption{Introduction of problem scenarios}
\label{Introduction of problem scenario}
\setlength{\tabcolsep}{13.3pt}
\begin{tabular}{lllll}
\hline
Problems & $size$  & $n$   & $yield$ & $distance$ \\
\hline
1       & 20$\times$20 & 40  & 2099  & 19.0262  \\
2       & 20$\times$20 & 60  & 3295  & 21.0237  \\
3       & 20$\times$20 & 80  & 4334  & 21.0237  \\
4       & 30$\times$30 & 90  & 5014  & 31.3847  \\
5       & 30$\times$30 & 135 & 7519  & 32.2024  \\
6       & 30$\times$30 & 180 & 10236 & 32.2024  \\
7       & 40$\times$40 & 160 & 8732  & 42.5441  \\
8       & 40$\times$40 & 240 & 13211 & 43.3821  \\
9       & 40$\times$40 & 320 & 17629 & 43.3821  \\
10      & 50$\times$50 & 250 & 13663 & 54.5619  \\
11      & 50$\times$50 & 375 & 20752 & 54.5619  \\
12      & 50$\times$50 & 500 & 27266 & 54.5619  \\
13      & 60$\times$60 & 360 & 19980 & 65.7419  \\
14      & 60$\times$60 & 540 & 29819 & 65.7419  \\
15      & 60$\times$60 & 720 & 39816 & 65.7419 \\
\hline
\end{tabular}
\end{table}

\subsection{Experimental setup}
\label{Experimental setup}
To thoroughly evaluate HRRA's performance, $15$ benchmark problems are developed with varying complexity levels. As detailed in Table~\ref{Introduction of problem scenario}, these problems differ significantly in their scenario size ($size$), number of tasks ($n$), total yield ($yield$), and maximum distance ($distance$) between task locations and depot. The yield at each task location is randomly generated within the interval [40, 70]. With a consistent population size of 30 and the number of robots set to \{4, 5, 6\}~\cite{dai2023multi}, $45$ test instances are constructed.

Algorithm performance is assessed using both modified inverted generational distance (IGD$^+$)~\cite{audet2021performance} and hypervolume (HV)~\cite{shang2020survey} metrics. The IGD$^+$ calculation utilizes reference points derived from the approximate PF, which is constructed through linear interpolation~\cite{blu2004linear} of non-dominated solutions obtained from 10 independent runs of all compared algorithms in this study. A smaller IGD$^+$ value indicates better solution quality, as it represents smaller average distances from these reference points to the obtained solution set, where distances are calculated to penalize only the objective components in which solutions fail to meet or outperform the reference points. Conversely, the HV metric, whose reference point is (1,1), favors solutions that maximize the dominated hypervolume, with larger values indicating superior performance. Since only non-dominated solutions provide meaningful insights for decision-makers in the AMERTA problem, our evaluation focuses exclusively on the non-dominated solutions within each algorithm's final population.

To ensure fair comparison, all algorithms are terminated based on CPU time limit of 0.5$\times n$ seconds. Experiments are conducted in MATLAB 2021a on a computing platform equipped with an Intel Core i7-12700 CPU (2.1 GHz, 2.1 GHz) and 32GB RAM.

\begin{table}[!htbp]
\centering
\caption{Summary of comparison results with Wilcoxon test}
\label{Comparison results of each algorithm on different problems}
\setlength{\tabcolsep}{6pt} 
\centering\small
\scalebox{0.685}{ 
\begin{tabular}{@{}lcccccccc@{}}
\hline
\textbf{HRRA VS.} & \multicolumn{7}{c}{\textbf{IGD$^+$}} \\
\cline{2-8} 
\textbf{(+/-/=)} & \textbf{AMOEA} & \textbf{CDABC} & \textbf{MODABC} & \textbf{NSGA-\uppercase\expandafter{\romannumeral2}} & \textbf{RNSGA} & \textbf{IALNS} & \textbf{HACO} \\ 
\hline
\textit{r}=4 & 0/15/0 & 0/12/3 & 0/13/2 & 0/15/0 & 1/9/5 & 1/12/2 & 1/8/6 \\
\textit{r}=5 & 0/15/0 & 0/13/2 & 0/13/2 & 0/15/0 & 1/13/1 & 1/11/3 & 3/10/2 \\
\textit{r}=6 & 0/15/0 & 0/14/1 & 0/14/1 & 0/15/0 & 0/14/1 & 0/11/4 & 0/6/9 \\ 
\hline
\textbf{HRRA VS.} & \multicolumn{7}{c}{\textbf{HV}} \\
\cline{2-8} 
\textbf{(+/-/=)} & \textbf{AMOEA} & \textbf{CDABC} & \textbf{MODABC} & \textbf{NSGA-\uppercase\expandafter{\romannumeral2}} & \textbf{RNSGA} & \textbf{IALNS} & \textbf{HACO} \\ 
\hline
\textit{r}=4 & 0/15/0 & 0/14/1 & 0/15/0 & 0/15/0 & 0/13/2  & 0/13/2 & 0/14/1\\
\textit{r}=5 & 0/15/0 & 0/15/0 & 0/15/0 & 0/15/0 & 0/13/2 & 0/12/3 & 0/14/1 \\
\textit{r}=6 & 0/15/0 & 0/15/0 & 0/15/0 & 0/15/0 & 0/15/0  & 0/11/4 & 0/14/1\\ 
\hline
\end{tabular}
}
\end{table}

\begin{figure*}[!htbp]
    \centering
    \begin{subfigure}
        \centering
        \includegraphics[width=\textwidth]{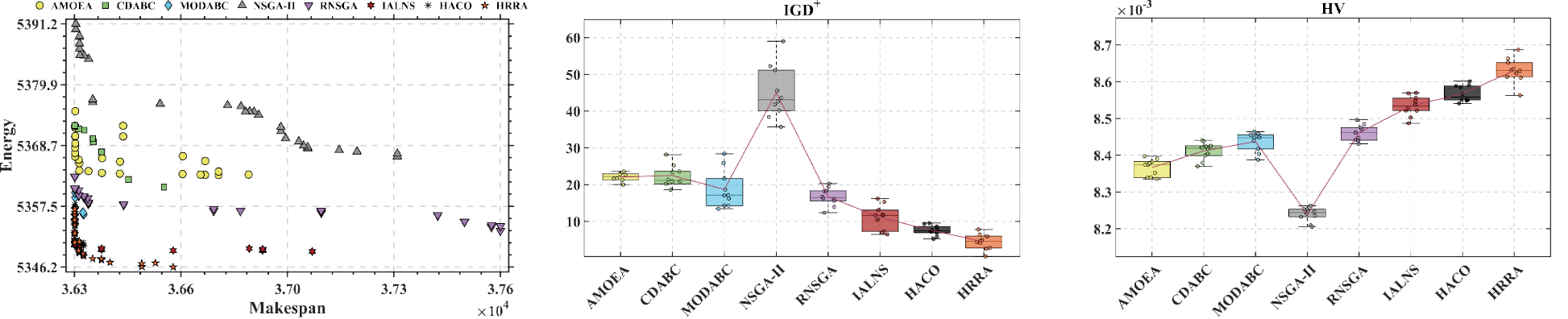}
        \caption*{(a) Problem 6 with $r = 4$}
        \label{fig:pf_a}
    \end{subfigure}
    \vspace{2mm}
    
    \begin{subfigure}
        \centering
        \includegraphics[width=\textwidth]{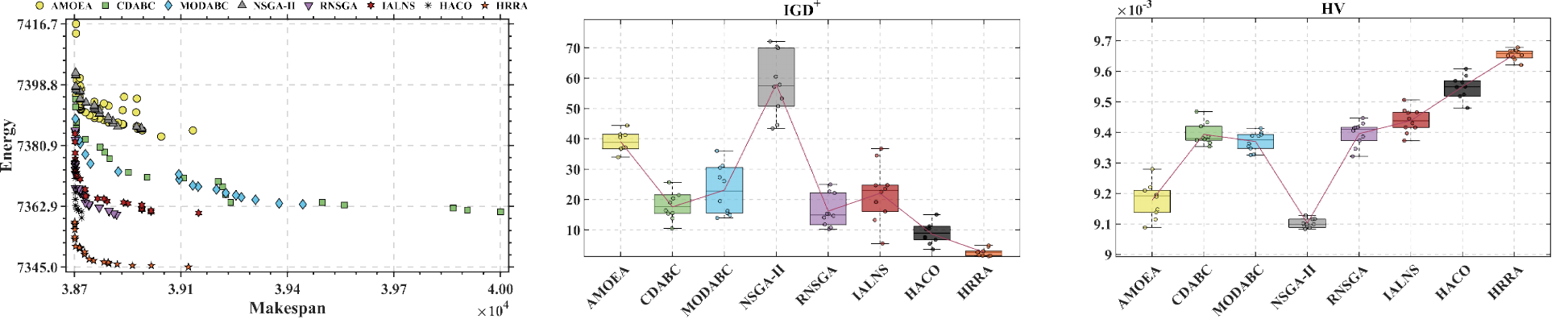}
        \caption*{(b) Problem 10 with $r = 5$}
        \label{fig:pf_b}
    \end{subfigure}
    \vspace{2mm}
    
    \begin{subfigure}
        \centering
        \includegraphics[width=\textwidth]{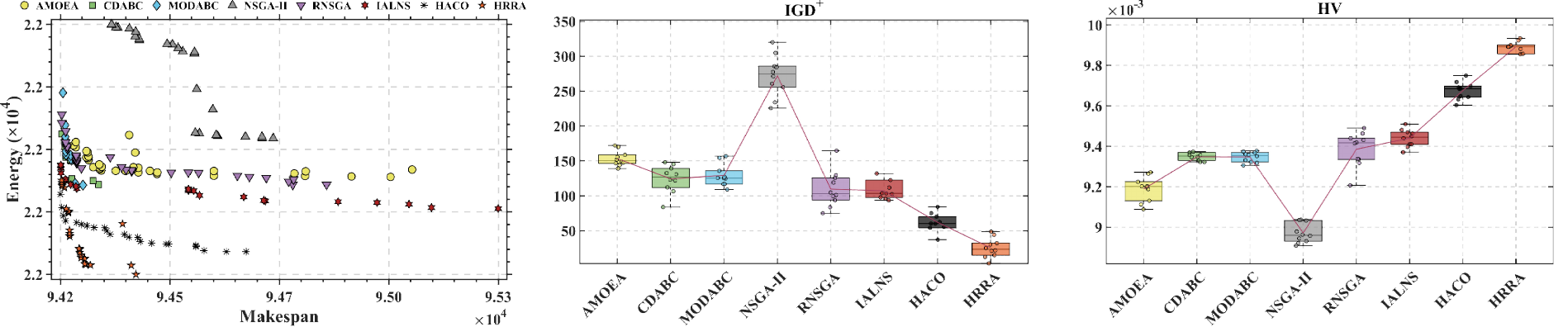}
        \caption*{(c) Problem 15 with $r = 6$}
        \label{fig:pf_c}
    \end{subfigure}
    
    \caption{Results obtained by compared algorithms on representative problem instances}
    \label{fig:pf}
\end{figure*}

\subsection{Comparative experiments and analysis}
\label{sec:experiments }

\subsubsection{Experimental results presentation and analysis}

The proposed HRRA algorithm undergoes comprehensive evaluation against seven representative benchmark algorithms: AMOEA~\cite{dong2024effective}, CDABC~\cite{guo2024effective}, MODABC~\cite{dai2023multi}, NSGA-\uppercase\expandafter{\romannumeral2}~\cite{deb2002fast}, RNSGA~\cite{chen2025reinforced}, IALNS~\cite{amiri2023robust}, and HACO~\cite{comert2023new}. The first three algorithms specifically address agricultural MRTA problems, while RNSGA is an approach for the general MRTA problem that employs a hierarchical hybrid encoding structure. IALNS and HACO originate as weighted single-objective methods for multi-objective EVRP. To enhance their multi-objective capability, we incorporate a non-dominated sorting mechanism. NSGA-\uppercase\expandafter{\romannumeral2} is a classic benchmark algorithm for comparison in both agricultural MRTA and EVRP. Therefore, we adopt an improved version described in~\cite{dai2023multi} to strengthen its combinatorial optimization performance.

Table~\ref{Comparison results of each algorithm on different problems} summarizes the experimental results. Symbolic annotations (`+', `-', `=') denote statistically significant superiority, inferiority, or equivalence relative to HRRA~\cite{chen2024archive}. Detailed results with $r = 4$ are presented in Table~\ref{resultsr=4}; the remaining results are shown in Tables~\ref{resultsr=5} - \ref{resultsr=6} in Section \ref{S-table} due to page limitations. The overall statistical analysis reveals HRRA's superior performance, with lower average IGD$^+$ values in $71.1\%$ of test instances and higher average HV values in $93.3\%$ of cases compared to other algorithms. This quantitative evidence is further strengthened by the PF and boxplot analysis in Figs.~\ref{fig:r4_1} - \ref{fig:r6_1} in Section \ref{S-figures}. Fig.~\ref{fig:pf} demonstrates the performance differences among algorithms through representative test scenarios. The PF distributions visually confirm that HRRA's solution set occupies the most advanced positions in the objective space. Boxplots extensively reveal that HRRA not only achieves superior mean values in both IGD$^+$ and HV metrics but also exhibits smaller interquartile ranges, demonstrating enhanced stability and robustness.

Through in-depth analysis, HRRA's performance exhibits a strong correlation with problem scale and complexity. In small-scale scenarios with robot redundancy (Fig.~\ref{fig:pf}(a)), the CRRM demonstrates a limited contribution due to reduced charging demands, even introducing computational overhead. However, in scenarios with high robot utilization (Fig.~\ref{fig:pf}(b)), the SRRM effectively optimizes makespan through adaptive path segmentation and task reallocation. As problem complexity increases (Fig.~\ref{fig:pf}(c)), HRRA's comprehensive advantages become increasingly pronounced.

\begin{table*}[htbp]
\centering
\caption{Comparison results of each algorithm with Wilcoxon test on different problems with $r=4$}
\label{resultsr=4}
\setlength{\tabcolsep}{6.4pt}
\renewcommand{\arraystretch}{1.2} 
\centering\small
\scalebox{0.58}{
\begin{tabular}{cllllllll}
\hline
\multicolumn{9}{c}{IGD$^+$}                                                                                                                                                                                                           \\
\hline
Problem & \multicolumn{1}{c}{AMOEA} & \multicolumn{1}{c}{CDABC} & \multicolumn{1}{c}{MODABC} & \multicolumn{1}{c}{NSGA-\uppercase\expandafter{\romannumeral2}} & \multicolumn{1}{c}{RNSGA} & \multicolumn{1}{c}{IALNS} & \multicolumn{1}{c}{HACO} & \multicolumn{1}{c}{HRRA} \\
\hline
1       & 2.7175e+00 (6.65e-01) - & 1.2838e+00 (3.24e-01) - & 1.4816e+00 (3.72e-01) - & 3.9895e+00 (6.51e-01) - & 8.1749e-01 (2.32e-01) =                                  & 1.2525e+00 (4.82e-01) -                                  & 1.4783e+00 (4.23e-01) -                                  & \cellcolor[HTML]{D9D9D9}\textbf{6.3046e-01 (1.79e-01)} \\
2       & 6.2314e+00 (8.55e+00) - & 1.8713e+00 (4.54e-01) - & 1.6661e+00 (8.88e-01) = & 3.5607e+00 (7.05e-01) - & 1.2364e+02 (7.74e-02) -                                  & 1.8453e+00 (5.05e-01) -                                  & 1.1841e+00 (5.10e-01) =                                  & \cellcolor[HTML]{D9D9D9}\textbf{8.8825e-01 (7.28e-01)} \\
3       & 4.0931e+00 (9.74e-01) - & 1.9638e+00 (1.09e+00) = & 2.0812e+00 (9.66e-01) - & 5.8285e+00 (7.48e-01) - & \cellcolor[HTML]{D9D9D9}\textbf{7.5841e-01 (3.85e-01) =} & 1.2444e+00 (7.29e-01) =                                  & 1.4128e+00 (1.13e+00) =                                  & 1.1965e+00 (6.41e-01)                                  \\
4       & 2.9740e+00 (1.15e+00) - & 3.5370e+00 (2.01e+00) - & 3.9245e+00 (1.36e+00) - & 1.1994e+01 (2.82e+00) - & 1.7861e+00 (9.26e-01) =                                  & 1.9729e+00 (4.65e-01) -                                  & 2.0746e+00 (1.20e+00) =                                  & \cellcolor[HTML]{D9D9D9}\textbf{1.2516e+00 (6.56e-01)} \\
5       & 6.4279e+00 (2.33e+00) - & 4.6935e+00 (2.10e+00) - & 5.6360e+00 (2.54e+00) - & 2.5204e+01 (6.12e+00) - & \cellcolor[HTML]{D9D9D9}\textbf{2.0015e+00 (9.31e-01) =} & 2.8045e+00 (1.21e+00) =                                  & 2.5032e+00 (2.39e+00) =                                  & 2.5052e+00 (1.27e+00)                                  \\
6       & 2.2403e+01 (2.79e+00) - & 2.1665e+01 (3.78e+00) - & 1.8720e+01 (5.08e+00) - & 4.5014e+01 (7.13e+00) - & 1.6738e+01 (2.44e+00) -                                  & 1.1100e+01 (3.38e+00) -                                  & 7.5619e+00 (1.38e+00) -                                  & \cellcolor[HTML]{D9D9D9}\textbf{4.4494e+00 (2.10e+00)} \\
7       & 1.7108e+01 (1.79e+00) - & 6.0715e+00 (3.31e+00) - & 5.1860e+00 (2.41e+00) - & 1.7460e+01 (4.00e+00) - & 2.8409e+00 (2.57e+00) =                                  & 6.5488e+00 (2.72e+00) -                                  & 5.0538e+00 (2.19e+00) -                                  & \cellcolor[HTML]{D9D9D9}\textbf{2.3273e+00 (2.48e+00)} \\
8       & 3.4996e+01 (4.14e+00) - & 1.6339e+01 (4.99e+00) - & 1.6350e+01 (4.27e+00) - & 3.7227e+01 (1.88e+00) - & 1.4176e+01 (4.69e+00) -                                  & 7.0607e+00 (2.27e+00) -                                  & 7.5778e+00 (1.55e+00) -                                  & \cellcolor[HTML]{D9D9D9}\textbf{1.3203e+00 (7.21e-01)} \\
9       & 4.4372e+01 (3.78e+00) - & 2.7754e+01 (5.95e+00) - & 2.4783e+01 (5.86e+00) - & 5.8091e+01 (4.57e+00) - & 2.2426e+01 (3.13e+00) -                                  & 2.8327e+01 (5.55e+00) -                                  & 1.0770e+01 (4.92e+00) -                                  & \cellcolor[HTML]{D9D9D9}\textbf{6.0394e+00 (2.06e+00)} \\
10      & 4.0581e+01 (6.29e+00) - & 2.6812e+01 (6.10e+00) - & 3.1447e+01 (7.10e+00) - & 6.4448e+01 (1.08e+01) - & 2.1813e+01 (4.85e+00) -                                  & 9.1825e+00 (4.24e+00) -                                  & 1.1808e+01 (7.06e+00) -                                  & \cellcolor[HTML]{D9D9D9}\textbf{1.9363e+00 (1.33e+00)} \\
11      & 5.8087e+01 (6.02e+00) - & 5.2653e+01 (7.99e+00) - & 5.1473e+01 (4.29e+00) - & 1.0174e+02 (4.90e+00) - & 4.6271e+01 (5.82e+00) -                                  & 4.9564e+01 (1.70e+01) -                                  & 2.7571e+01 (4.47e+00) -                                  & \cellcolor[HTML]{D9D9D9}\textbf{1.0880e+01 (6.02e+00)} \\
12      & 8.5821e+01 (5.41e+00) - & 6.9854e+01 (1.34e+01) - & 6.7865e+01 (1.83e+01) - & 1.5577e+02 (2.83e+01) - & 7.4533e+01 (1.60e+01) -                                  & 5.2417e+01 (1.57e+01) -                                  & 1.9079e+01 (6.13e+00) =                                  & \cellcolor[HTML]{D9D9D9}\textbf{1.6078e+01 (6.11e+00)} \\
13      & 7.8357e+01 (1.11e+01) - & 5.5907e+01 (1.48e+01) - & 6.4423e+01 (1.20e+01) - & 9.8441e+01 (6.98e+00) - & 4.6429e+01 (1.24e+01) -                                  & 2.7179e+01 (9.54e+00) -                                  & 2.3596e+01 (1.16e+01) -                                  & \cellcolor[HTML]{D9D9D9}\textbf{4.3310e+00 (2.90e+00)} \\
14      & 8.8906e+01 (9.34e+00) - & 5.0656e+01 (1.16e+01) = & 5.2854e+01 (1.53e+01) = & 1.1596e+02 (9.18e+00) - & 3.7683e+01 (6.16e+00) +                                  & \cellcolor[HTML]{D9D9D9}\textbf{1.3801e+01 (4.73e+00) +} & 2.3163e+01 (1.64e+01) +                                  & 4.8711e+01 (7.39e+00)                                  \\
15      & 1.1248e+02 (1.18e+01) - & 7.7822e+01 (2.31e+01) = & 8.1320e+01 (1.54e+01) - & 1.9126e+02 (4.04e+01) - & 1.1807e+02 (7.52e+00) -                                  & 1.0179e+02 (3.79e+01) -                                  & \cellcolor[HTML]{D9D9D9}\textbf{4.2476e+01 (2.19e+01) =} & 5.8237e+01 (2.24e+01)                                  \\
\hline
\multicolumn{1}{c}{+/-/=} & \multicolumn{1}{c}{0/15/0} & \multicolumn{1}{c}{0/12/3} & \multicolumn{1}{c}{0/13/2} & \multicolumn{1}{c}{0/15/0} & \multicolumn{1}{c}{1/9/5} & \multicolumn{1}{c}{1/12/2} & \multicolumn{1}{c}{1/8/6} & \\
\hline
\multicolumn{9}{c}{HV}                                                                                                                                                                                                                           \\
\hline
Problem & \multicolumn{1}{c}{AMOEA} & \multicolumn{1}{c}{CDABC} & \multicolumn{1}{c}{MODABC} & \multicolumn{1}{c}{NSGA-\uppercase\expandafter{\romannumeral2}} & \multicolumn{1}{c}{RNSGA} & \multicolumn{1}{c}{IALNS} & \multicolumn{1}{c}{HACO} & \multicolumn{1}{c}{HRRA} \\
\hline
1       & 1.4652e-02 (8.58e-05) - & 1.4791e-02 (4.32e-05) - & 1.4765e-02 (4.01e-05) - & 1.4606e-02 (7.22e-05) - & 1.4847e-02 (1.94e-05) -                                  & 1.4783e-02 (4.29e-05) -                                  & 1.4790e-02 (5.07e-05) -                                  & \cellcolor[HTML]{D9D9D9}\textbf{1.4895e-02 (3.43e-05)} \\
2       & 9.3110e-03 (9.87e-05) - & 9.4165e-03 (2.43e-05) - & 9.4134e-03 (4.13e-05) - & 9.3129e-03 (2.04e-05) - & 8.3982e-03 (1.90e-05) -                                  & 9.4363e-03 (3.14e-05) -                                  & 9.4104e-03 (3.46e-05) -                                  & \cellcolor[HTML]{D9D9D9}\textbf{9.4771e-03 (3.06e-05)} \\
3       & 1.0587e-02 (2.06e-05) - & 1.0655e-02 (3.42e-05) = & 1.0620e-02 (3.91e-05) - & 1.0487e-02 (2.40e-05) - & \cellcolor[HTML]{D9D9D9}\textbf{1.0690e-02 (2.62e-05) =} & 1.0669e-02 (5.66e-05) =                                  & 1.0659e-02 (2.87e-05) =                                  & 1.0682e-02 (2.14e-05)                                  \\
4       & 1.4094e-02 (5.35e-05) - & 1.4085e-02 (6.90e-05) - & 1.4074e-02 (4.58e-05) - & 1.3901e-02 (5.16e-05) - & 1.4134e-02 (5.89e-05) -                                  & 1.4107e-02 (7.10e-05) -                                  & 1.4123e-02 (4.58e-05) -                                  & \cellcolor[HTML]{D9D9D9}\textbf{1.4215e-02 (3.01e-05)} \\
5       & 1.1593e-02 (4.68e-05) - & 1.1599e-02 (5.64e-05) - & 1.1597e-02 (6.22e-05) - & 1.1293e-02 (4.11e-05) - & 1.1676e-02 (3.22e-05) =                                  & 1.1653e-02 (8.23e-05) =                                  & 1.1649e-02 (3.60e-05) -                                  & \cellcolor[HTML]{D9D9D9}\textbf{1.1709e-02 (2.93e-05)} \\
6       & 8.3660e-03 (2.29e-05) - & 8.4119e-03 (2.36e-05) - & 8.4373e-03 (2.57e-05) - & 8.2388e-03 (1.88e-05) - & 8.4610e-03 (2.11e-05) -                                  & 8.5673e-03 (2.21e-05) -                                  & 8.5340e-03 (2.72e-05) -                                  & \cellcolor[HTML]{D9D9D9}\textbf{8.6308e-03 (3.34e-05)} \\
7       & 1.2785e-02 (5.64e-05) - & 1.3049e-02 (6.41e-05) - & 1.3104e-02 (4.16e-05) - & 1.2824e-02 (7.89e-05) - & 1.3135e-02 (6.17e-05) -                                  & 1.3103e-02 (4.97e-05) -                                  & 1.3061e-02 (5.66e-05) -                                  & \cellcolor[HTML]{D9D9D9}\textbf{1.3205e-02 (2.51e-05)} \\
8       & 8.4271e-03 (4.14e-05) - & 8.6238e-03 (4.58e-05) - & 8.6181e-03 (2.96e-05) - & 8.3999e-03 (3.00e-05) - & 8.6504e-03 (3.33e-05) -                                  & 8.7362e-03 (2.61e-05) -                                  & 8.7375e-03 (2.72e-05) -                                  & \cellcolor[HTML]{D9D9D9}\textbf{8.8666e-03 (1.43e-05)} \\
9       & 9.2876e-03 (2.90e-05) - & 9.4407e-03 (5.28e-05) - & 9.4529e-03 (5.25e-05) - & 9.2268e-03 (3.89e-05) - & 9.4778e-03 (2.90e-05) -                                  & 9.6126e-03 (4.39e-05) -                                  & 9.4882e-03 (3.20e-05) -                                  & \cellcolor[HTML]{D9D9D9}\textbf{9.7070e-03 (2.89e-05)} \\
10      & 9.3948e-03 (4.15e-05) - & 9.5451e-03 (4.99e-05) - & 9.5233e-03 (5.07e-05) - & 9.1584e-03 (6.60e-05) - & 9.5981e-03 (4.89e-05) -                                  & 9.7549e-03 (6.25e-05) -                                  & 9.7410e-03 (5.97e-05) -                                  & \cellcolor[HTML]{D9D9D9}\textbf{9.8812e-03 (2.23e-05)} \\
11      & 8.0537e-03 (4.38e-05) - & 8.0754e-03 (5.32e-05) - & 8.0855e-03 (3.45e-05) - & 7.7391e-03 (2.52e-05) - & 8.1214e-03 (3.73e-05) -                                  & 8.2710e-03 (2.05e-05) -                                  & 8.1345e-03 (7.64e-05) -                                  & \cellcolor[HTML]{D9D9D9}\textbf{8.4065e-03 (4.43e-05)} \\
12      & 8.8624e-03 (3.72e-05) - & 8.9968e-03 (5.70e-05) - & 8.9931e-03 (8.52e-05) - & 8.6880e-03 (3.98e-05) - & 8.9960e-03 (6.12e-05) -                                  & 9.2798e-03 (3.53e-05) -                                  & 9.0995e-03 (6.40e-05) -                                  & \cellcolor[HTML]{D9D9D9}\textbf{9.4400e-03 (1.66e-05)} \\
13      & 8.5735e-03 (8.02e-05) - & 8.8129e-03 (4.79e-05) - & 8.7804e-03 (5.63e-05) - & 8.5133e-03 (3.23e-05) - & 8.8497e-03 (5.54e-05) -                                  & 9.0014e-03 (5.79e-05) -                                  & 8.9775e-03 (5.08e-05) -                                  & \cellcolor[HTML]{D9D9D9}\textbf{9.1760e-03 (2.85e-05)} \\
14      & 8.4016e-03 (4.34e-05) - & 8.5909e-03 (4.46e-05) - & 8.5924e-03 (5.24e-05) - & 8.3395e-03 (3.19e-05) - & 8.6470e-03 (3.08e-05) -                                  & 8.7469e-03 (5.50e-05) -                                  & 8.7779e-03 (2.95e-05) -                                  & \cellcolor[HTML]{D9D9D9}\textbf{8.8842e-03 (1.87e-05)} \\
15      & 8.4961e-03 (3.33e-05) - & 8.6682e-03 (4.51e-05) - & 8.6440e-03 (3.67e-05) - & 8.4294e-03 (3.77e-05) - & 8.6701e-03 (5.45e-05) -                                  & 8.8367e-03 (6.51e-05) -                                  & 8.6512e-03 (5.55e-05) -                                  & \cellcolor[HTML]{D9D9D9}\textbf{9.0281e-03 (2.30e-05)} \\
\hline
\multicolumn{1}{c}{+/-/=} & \multicolumn{1}{c}{0/15/0} & \multicolumn{1}{c}{0/14/1} & \multicolumn{1}{c}{0/15/0} & \multicolumn{1}{c}{0/15/0} & \multicolumn{1}{c}{0/13/2} & \multicolumn{1}{c}{0/13/2} & \multicolumn{1}{c}{0/14/1} & \\\hline
\end{tabular}
}
\end{table*}

\begin{figure}[htp]
    \centering
    \includegraphics[width=8.8cm]{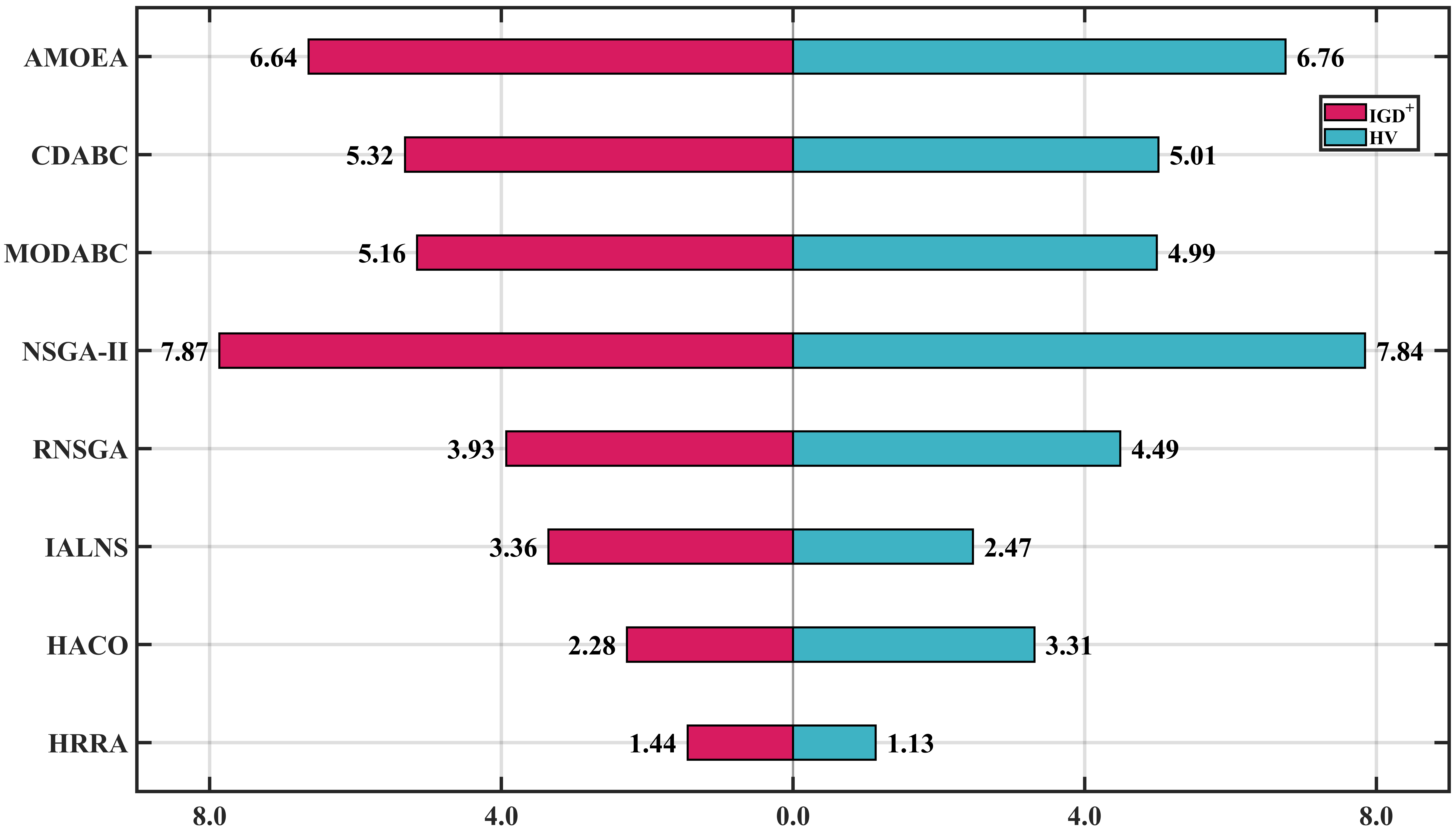}
    \caption{Comparison of algorithm rankings obtained by the Friedman test}
    \label{fig:friedman}
\end{figure}

Comprehensive Wilcoxon signed-rank test results~\cite{derrac2011practical} across the 45 test instances in Table~\ref{wilxocon} reveal that, for both the IGD$^+$ and HV metrics, the R$^+$ values for HRRA are substantially greater than the R$^-$ values when compared against each competitor, indicating that HRRA statistically significantly outperforms these algorithms. All the associated $P$-values $<0.05$ further affirm that these observed performance advantages are statistically significant and not attributable to random chance. Moreover, HRRA consistently achieved the foremost rank in the Friedman test in Fig.~\ref{fig:friedman}, which further corroborates this conclusion of its overall superiority. This collective statistical evidence strongly supports HRRA's capacity to deliver robustly superior and consistent optimization performance across diverse problem instances.

\begin{table}[htbp]
  \centering
  \caption{Wilcoxon signed-rank test results between HRRA and the compared algorithms}
  \setlength{\tabcolsep}{13pt}
\centering\small
\begin{tabular}{cccc}
\hline
\multicolumn{4}{c}{{IGD$^+$}}\\
\hline
 HRRA VS. & $R^{+}$ & $R^{-}$  &  $P-$value $\le$ 0.05 \\ \hline 
AMOEA & 1035.0 & 0.0  & Yes\\ \hline 
CDABC & 1035.0 & 0.0  & Yes\\ \hline 
MODABC & 1023.0 & 12.0  & Yes\\ \hline 
NSGA-\uppercase\expandafter{\romannumeral2} & 1035.0 & 0.0  & Yes\\ \hline 
RNSGA & 998.5 & 36.5  & Yes\\ \hline 
IALNS & 945.0 & 90.0  & Yes\\ \hline 
HACO & 754.0 & 281.0  & Yes\\ \hline 
    \multicolumn{4}{c}{{HV}}\\
    \hline
 HRRA VS. & $R^{+}$ & $R^{-}$  &  $P-$value $\le$ 0.05 \\ \hline 
AMOEA & 1035.0 & 0.0 & Yes \\ \hline 
CDABC & 990.0 & 0.0 & Yes \\ \hline 
MODABC & 1035.0 & 0.0 & Yes \\ \hline 
NSGA-\uppercase\expandafter{\romannumeral2} & 1035.0 & 0.0 & Yes\\ \hline 
RNSGA & 1033.5 & 1.5 & Yes\\ \hline 
IALNS & 965.5 & 24.5  & Yes\\ \hline 
HACO & 1033.5 & 1.5  & Yes\\ \hline 
 \hline
    \end{tabular}%
  \label{wilxocon}%
\end{table}%

\subsubsection{Performance attribution analysis}

MODABC pioneers agricultural MRTA optimization through a tri-phase search strategy (employed bee, onlooker bee, and scout bee) guided by an experience archive for local search operator selection. However, insufficient phase coordination and excessive randomness in the selection of tasks to be optimized constrain its performance stability. CDABC improves MODABC through deep local search for the most energy-consuming robots but exhibits inadequate workload balancing for makespan optimization. AMOEA emphasizes task balancing between the robots with the largest and smallest workloads, yet its over-concentrated local search resource allocation paradoxically restricts global optimization. Crucially, all three algorithms utilize the conventional encoding scheme that constrains intra-robot route optimization.

NSGA-\uppercase\expandafter{\romannumeral2} demonstrates robust performance in general multi-objective optimization through fast non-dominated sorting and crowding distance-based diversity maintenance. However, its ordinary genetic operators struggle to maintain solution feasibility in complex combinatorial spaces. The absence of targeted route local search mechanisms further restricts its capability for robot path and task sequence optimization. 

RNSGA employs a hierarchical and hybrid encoding of solutions, which facilitates a multi-level optimization process. This structure allows individual robot routes to be optimized locally, while the combined routes can be optimized from a global perspective. However, it lacks specialized mechanisms to proactively manage the complex battery constraints inherent in the AMERTA problem, potentially leading to suboptimal energy management strategies.

IALNS builds upon the neighborhood search framework and utilizes historical operator success to control the usage probabilities of different operators. Additionally, it integrates a simulated annealing acceptance criterion to manage the acceptance of intermediate non-improving solutions. Nevertheless, as its search is fundamentally guided by a single solution, it is inherently more susceptible to premature convergence.

HACO establishes a two-stage optimization framework by integrating ACO and ABC algorithms. While the ACO phase generates high-quality initial solutions with low computational resource consumption (validated by its strong performance in complex problems), the ABC phase demonstrates insufficient coordination in MRTA despite enhanced single-robot energy optimization. Additionally, its solution encoding scheme also limits deep optimization of individual robot routes. 

In contrast, HRRA's superiority manifests through three key innovations: First, its novel encoding scheme enables hierarchical local search across routes, robot-task mapping sequence, and global task sequences. Second, the CRRM and SRRM achieve precise optimization of specific task groups while preserving existing optimization results, effectively improving critical performance metrics. The absence of CRRM leads to inefficient energy management, underscoring its importance in balancing charging demands with task efficiency, while the absence of SRRM eliminates the algorithm's ability for fine-grained path optimization. Third, the synergistic operation of these mechanisms ensures algorithmic stability and superiority in complex problem scenarios.

\section{Conclusions and future work}
\label{conclusion}
This paper presents a novel hybrid algorithm for addressing an AMERTA problem. Through systematic theoretical analysis and experimental validation, this work yields several significant conclusions:

First, from a modeling perspective, this study pioneers the integration of load-dependent velocity variations and battery capacity constraints into agricultural MRTA problems. This establishes the AMERTA model, which better reflects real-world scenarios. This enhanced framework not only incorporates the traditional makespan-energy trade-off but also introduces complex constraints, providing a more comprehensive problem description for future research.

Second, regarding algorithmic design, HRRA achieves deep optimization over different levels through its hierarchical encoding structure. Compared to existing approaches, HRRA demonstrates notable advantages in several aspects: (1) the hierarchical optimization strategy enables simultaneous global exploration and local refinement while maintaining search efficiency; (2) the variable load-limit dual-phase initialization method effectively balances solution quality and diversity; (3) two optimization mechanisms enhance the efficiency of task sequencing; (4) the synergistic effect of CRRM and SRRM significantly enhances algorithm performance in complex scenarios, with correlation analysis confirming its unique search characteristics.

Third, experimental validation across 45 test instances of varying scales demonstrates HRRA's superior performance in both IGD$^+$ and HV metrics, particularly in large-scale complex problems. These results not only validate the algorithm's stability and robustness but also substantiate its potential for practical applications.

Building upon the current findings, multiple promising research directions warrant attention:
\begin{itemize}
\item Dynamic scenario adaptation: future research could extend to dynamic scenarios incorporating real-time task arrivals, robot failures, and real-time monitoring of task energy consumption, necessitating the development of online optimization mechanisms and rapid response strategies;

\item Integration of heterogeneous robot teams: future research could focus on incorporating heterogeneous robots (distinguished by varying attributes such as load capacities, speeds, energy consumption models, and operational capabilities) into the AMERTA problem model;


\item Cross-scenario applications: exploration of the algorithm's potential in analogous domains (e.g., warehouse logistics, urban distribution) would validate the model's transferability and algorithmic adaptability.
\end{itemize}

These research directions will not only enhance the algorithm's practicality but also advance the theoretical foundations of agricultural robot cooperation. As relevant technologies continue to evolve, we anticipate that HRRA-based optimization methods will play an increasingly significant role in smart agriculture applications.

\section*{Supplementary materials}

The supplementary materials for HRRA includes:
\begin{itemize}
    \item Section \ref{S-complexity}: algorithm complexity analysis;

    \item Section \ref{S-parameter}: parameter sensitivity analysis;

    \item Section \ref{S-ablation}: ablation study;

    \item Section \ref{S-default outputs}: performance comparison of the default outputs;

    \item Section \ref{S-Algorithms}: presentation of Algorithms~\ref{alg:initialization}~-~\ref{alg:SRRM};

    \item Section \ref{S-table}: detailed comparative results on the test instances in Tables~\ref{resultsr=5}~-~\ref{resultsr=6};

    \item Section \ref{S-figures}: visual presentation of all comparative results on the test instances in Figs.~\ref{fig:r4_1}~-~\ref{fig:r6_1}.
\end{itemize}










\bibliographystyle{unsrt}
\bibliography{ref}

\end{document}